\documentclass[a4paper, 12pt]{elsarticle}
\usepackage{enumerate}
\usepackage{lineno,hyperref}
\usepackage{lineno,hyperref,tikz,adjustbox}

\usepackage{mathtools}
\usepackage{algorithm}
\usepackage{algpseudocode}
\usepackage[margin=2.5cm]{geometry}

\usepackage{tikz-cd}
\tikzcdset{scale cd/.style={every label/.append style={scale=#1},
		cells={nodes={scale=#1}}}}

\modulolinenumbers[0]

\journal{Journal of \LaTeX\ Templates}

\usepackage{amssymb}
\usepackage{amsmath}
\usepackage{multirow}
\usepackage{longtable}
\usepackage{comment}
\usepackage{placeins}
\usepackage{mathtools}
\usepackage{algorithm}
\usepackage{algpseudocode}
\usepackage{enumitem}
\usepackage[utf8]{inputenc}
\usepackage{booktabs}
\usepackage{array}

\usepackage[draft,nomargin,inline]{fixme}
\fxsetface{inline}{\itshape}
\fxsetface{env}{\itshape}
\fxusetheme{color}

\usepackage{url}
\urldef{\mailsa}\path|{djukanovic, raidl}@ac.tuwien.ac.at,|
\urldef{\mailsb}\path|christian.blum@iiia.csic.es|

\usepackage{tikz}
\usetikzlibrary{positioning}
\definecolor{canaryyellow}{rgb}{1.0, 0.94, 0.0}
\definecolor{brightgreen}{rgb}{0.4, 1.0, 0.0}
\definecolor{jazzberryjam}{rgb}{0.65, 0.04, 0.37}

\usepackage{bookmark}
\usepackage{hypcap} 
\evensidemargin\oddsidemargin
\usepackage{graphicx}
\usepackage{xcolor}

\usepackage{xspace}
\usepackage{color}
\usepackage{epsfig}
\usepackage{caption}
\usepackage{subcaption}
\usepackage{mathrsfs}
\usepackage{amssymb}
\usepackage{amsmath}
\usepackage{amsthm}

\usepackage{blindtext}

\algnewcommand\algorithmicforeach{\textbf{for each}}
\algdef{S}[FOR]{ForEach}[1]{\algorithmicforeach\ #1\ \algorithmicdo}

\begin{document}
	
	\begin{frontmatter}
		
		\title{Graph Protection under Multiple Simultaneous Attacks: A Heuristic Approach}

		\author[1]{Marko Djukanovi\'c\fnref{myfootnote}}
		\address[1]{ Faculty of Natural Sciences and Mathematics, University of Banja Luka, Bosnia and Herzegovina}
		\fntext[1]{Corresponding author}
		\ead[1]{marko.djukanovic@pmf.unibl.org}
		
		\author[2]{Stefan Kapunac} 
		\address[2]{ Faculty of Mathematics, University of Belgrade, Serbia}
		\ead[2]{stefan.kapunac@matf.bg.ac.rs}
		
		\author[2]{Aleksandar Kartelj} 
		\ead[3]{aleksandar.kartelj@matf.bg.ac.rs}
		
		\author[1]{Dragan Mati\'c} 
		\ead[4]{dragan.matic@pmf.unibl.org}   
		
		\begin{abstract}
			This work focuses on developing an effective meta-heuristic approach to protect against simultaneous attacks on nodes of a network modeled using a graph. Specifically, we focus on the $k$-strong Roman domination problem, a generalization of the well-known Roman domination problem on graphs. This general problem is about assigning integer weights to nodes that represent the number of field armies stationed at each node in order to satisfy the protection constraints while minimizing the total weights. These constraints concern the protection of a graph against any simultaneous attack consisting of $k \in \mathbb{N}$ nodes. An attack is considered repelled if each node labeled 0 can be defended by borrowing an army from one of its neighboring nodes, ensuring that the neighbor retains at least one army for self-defense. 
			The $k$-SRD problem has practical applications in various areas, such as developing counter-terrorism strategies or managing supply chain disruptions. The solution to this problem is notoriously difficult to find, as even checking the feasibility of the proposed solution requires an exponential number of steps. We propose a variable neighborhood search algorithm in which the feasibility of the solution is checked by introducing the concept of quasi-feasibility, which is realized by careful sampling within the set of all possible attacks. 
			Extensive experimental evaluations show the scalability and robustness of the proposed approach compared to the two exact approaches from the literature. Experiments are conducted with random networks from the literature and newly introduced random wireless networks as well as with real-world networks. A practical application scenario, using real-world networks, involves applying our approach to graphs extracted from GeoJSON files containing geographic features of hundreds of cities or larger regions.
		\end{abstract}
		
		
	\end{frontmatter}
	
	
	\section{Introduction}\label{sec:introduction}
	
	Domination problems on graphs represent one of the most frequently solved class of problems in the field of computer science, thoroughly investigated theoretically as well as computationally (see surveys~\cite{cockayne1978domination,henning2009survey, goddard2013independent} providing an overview on theoretical results, and \cite{lu2010survey,chang1998algorithmic} for an overview on algorithmic results). Given an input graph $G=(V, E)$, the minimum domination problem asks for finding a minimum cardinalty subset $S$ of nodes of graph $G$, so that any node from $V$ either belongs to $S$ or is adjacent to at least one node from $S$.
	Many variants of the minimum domination problem are proposed, corresponding to a  a wide range of practical  applications, e.g.\ in telecommunications~\cite{balasundaram2006graph}, scheduling problems~\cite{rao2021survey}, and detecting RNA motifs in biological networks. Many of them are related to well-known facility location problems~\cite{gupta2013domination}. In the course of this work, we pay particular attention to a problem belonging to class of the \emph{Roman domination problems} (RDPs); these problems are thoroughly solved in the literature from both perspectives, theoretically and practically, see~\cite{chellali2020roman,cockayne2004roman}. The basic RDP seeks for a positive integer node labeling with the minimum cumulative weight over all nodes of a graph, such that every node labeled with 0 must have at least one neighboring node labeled with 2. This problem can be seen by ensuring a network with minimum cost for each possible individual (node) attack. RDP and its variants are well solved practically in the literature, e.g. by means of an integer linear program~\cite{poureidi2023algorithmic,burger2013binary}, a genetic algorithm~\cite{khandelwal2020roman}, a variable neighborhood search~\cite{ivanovic2019variable}, etc. Practical applications of RDP have been found in the search for an optimal military strategy~\cite{henning2003defending}, network design of server placements~\cite{pagourtzis2002server}, to name a few.
	
	In this paper, we study a practically motivated variant of RDP where $k\in \mathbb{N}, k \geq 2$ attacks occur simultaneously on the nodes of a graph; the problem is called \emph{k-Strong Roman domination} ($k$-SRD) \emph{problem}~\cite{liu2020k}. It is similar to a basic RDP, but instead of a single node being attacked, any combination of $k\geq 2$ nodes can be attacked, and all of these attacks must be defended against by the respective neighboring nodes. In reality, a scenario may arise where defenders need to be prepared for multiple simultaneous attacks in order to secure more than one site at a time, e.g., when planning counter-terrorism strategies. Other applications of this problem can be found in disaster relief or supply chain disruptions~\cite{liu2020k}. Verifying the feasibility of the proposed solution requires an exponential number of steps. To develop an efficient algorithm, we have defined the concept of $k$-SRD quasi-feasibility.
	We highlight another optimization problem that also considers multiple simultaneous attacks on graphs, but under different constraints; it is the \emph{strong Roman domination} problem~\cite{alvarez2017strong}.

	Concerning  the literature on the $k$-SRD problem, there are, to our knowledge, two papers in the existing literature. One of the papers deals with theoretical aspects of the problem~\cite{nikolic2023theoretical}. The second paper~\cite{liu2020k} examines the problem from a practical perspective. In this study, the authors propose two exact algorithms: an integer linear programming model and, based on it, a Benders-based decomposition technique; the latter approach slightly outperforms the former in terms of the number of optimally solved instances. The number of variables of the ILP model is $O\left(\binom{n}{k} \cdot n^2\right)$, while the number of constraints is $O\left(\binom{n}{k} \cdot n\right)$; thus, the size of the model grows exponentially with the number of variables and constraints as the instance size increases. Therefore, the model requires enormous computational resources to be constructed for large problem instances. As an alternative, a Benders-based decomposition approach was proposed in the same paper.
	Besides all these facts, we noticed some other problems. First, the approaches were only tested on a set of randomly generated instances of limited size with an instance size of up to 30 (nodes) and up to $k=5$. Therefore, these graph sizes can hardly model a realistic attack scenario, indicating potential problems with the scalability of the approaches. Moreover, random graphs may be unsuitable for modelling real-world use cases. Secondly, as already mentioned, exact approaches are usually only suitable to a limited extent for solving such difficult problems. A reasonable way is to consider approximation search methods, and one of them is certainly meta-heuristics. The contributions of this work are as follows.
	
	\begin{itemize}
		\item To solve the $k$-SRD problem on larger instances and thus more realistic graphs, we propose a variable neighborhood search (VNS) approach~\cite{mladenovic1997variable}, the first known heuristic approach to solve this problem. This method is equipped with an efficient procedure that verifies solution feasibility  based on the introduced concept of quasi-feasibility. For that purpose, two strategies are used to defend against a set of (predefined) attacks: a deterministic one, which is used when the number of attacks is rather small, and a more robust stochastic one, which is based on a roulette selection to choose neighboring nodes as defenders of the attacked nodes.
		
		\item A greedy approach was proposed to quickly find a feasible solution of adequate quality. This method also supports the VNS approach as the solution derived by the greedy approach serves as a starting point for the VNS.
		
		\item We extend the current set of random instances by creating additional sets of instances with different distributions. These are constructed using the model \emph{Unit disc} model~\cite{clark1990unit}, which generates random geometric graphs commonly used to model wireless ad hoc networks. The purpose of these instances is to test the robustness of the literature approaches together with the two proposed approaches under more realistic graphs of different densities.
		
		\item A new application scenario for the $k$-SRD problem has been provided using a set of real instances. More specifically, this use case is about applying the $k$-SRD solution to graphs representing cities divided into districts or regions. The task is to select suitable locations for the construction of fire stations and to place a number of fire trucks to ensure the extinguishing of fire disasters that may occur simultaneously at $k$ locations.
		
		\item Our VNS approach shows its effectiveness by achieving solutions that are close to (or equal to) the quality of known optimal solutions for small instances. For medium to large instances, the two exact approaches from the literature are far from being applicable (except partially for the case of small $k=2$), since it is impossible to construct the model within a reasonable memory consumption or the given time limit. This was not the case with our VNS, which has greater robustness compared to the exact methods. Moreover, VNS significantly improves the initial solutions generated by our greedy approach, as demonstrated by appropriate statistical tests.
	\end{itemize}

	\subsection{Problem definition}\label{sec:problem-definition}
	
	Let $G=(V(G),E(G))$ be a simple undirected graph where $V(G)$ and $E(G)$ represent its nodes and edges, respectively. Let $n=|V(G)|$ be the number of nodes of the graph $G$, also called the size of the graph. With $d(v), v \in V(G)$ we denote the degree of the node $v$, and with $\Delta(G)= \max \{ d(v) \mid v \in V(G) \}$ the maximum degree of the graph $G$.
	Let $f \colon V(G) \mapsto \mathbb{N}_0$ map the nodes of the graph to the numbers (of armies) and let $\omega(f, G) = \sum_{v \in V(G) } f(v)$ be the sum of these numbers for the entire $G$. (We always consider $n\geqslant k$ throughout the paper; otherwise we arrive at the trivial solution of the problem.)
	
	For a graph $G$, let $P := \{v_1, v_2, \ldots, v_k\}$, $v_1, v_2, \ldots , v_k \in V(G)$ denote an attack pattern, or simply, the set of nodes under attack. Note that there are $\binom{|V(G)|}{k}$ different attacks for the graph $G$ and the fixed value of $k$. Therefore,  the number of attacks grows exponentially in the instance size ($n$). Let $\mathcal{P}(G):= \{ P_1, \ldots, P_{\binom{|V(G)|}{k}}\}$ be the set of enumerated attacks for graph $G$.
	
	A function $f: V(G) \mapsto \{0,1, \ldots, \min(\Delta(G), k) + 1\}$ is a proper $k$-SRD function if each attack $P \in \mathcal{P}(G)$ is defended, where the defense rules are as follows:
	\begin{itemize}
		\item [($i$)] Nodes labeled with $f(v)=1$ can defend only themselves. 
		\item [($ii$)] Nodes labeled with $f(v)\geq 2$ can defend themselves and, in addition, at most $f(v)-1$ attacked neighboring nodes labeled with 0.
		\item[($iii$)] Nodes labeled with 0 that are under attack $P$ must be defended by at least one of their neighboring nodes.
	\end{itemize}
	
\begin{figure}[!ht]
		\centering
		\begin{tikzpicture}
			\node[circle, draw, fill=gray] (A) at (0, 0) {1};
			\node[circle, draw] (B) at (2, 0) {0};
			\node[circle, draw, fill=darkgray] (C) at (1, 1.7) {2};
			\node[circle, draw, fill=gray] (D) at (1, 3.2) {1};
			\node[circle, draw, fill=gray] (E) at (4, 0) {1};
			\draw (A) -- (B);
			\draw (B) -- (C);
			\draw (C) -- (A);
			\draw (C) -- (D);
			\draw (B) -- (E);
		\end{tikzpicture}
		\caption{3-SRD problem: An optimal strategy for the $3$-SRD function $f$. It comes with $\omega(f)=5$.
		}
		\label{fig:example-optimal-strategy}
	\end{figure}
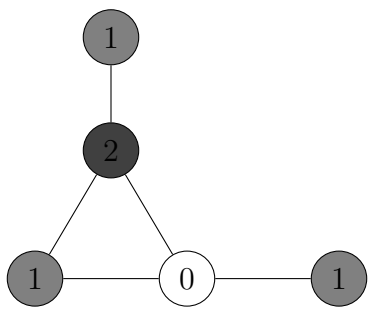

	The problem of $k$-SRD on the graph $G$ seeks for a valid $k$-SRD function $f$ on $G$ with minimum $\omega(f)$, called $k$-SRD number $\gamma_{k-SRD}(G)$, i.e. $\gamma_{k-SRD}(G):= \min\{ \omega(f) \mid f \text{ is a proper } k\text{-SRD function on the graph } G \}$. Such a $k$-SRD function for which $\gamma_{k-SRD}(G)$ is reached is simply called $\gamma_{k-SRD}$ function. Figure~\ref{fig:defended-attack} depicts a graph and an attack for which the defense strategy is given. Note that the solution $f$ is not a $3$-SRD function, since three more attacks need to be defended. For example, an attack $P=\{A,B, D\}$ could not be defended with $f$. An optimal strategy for defending against all attacks on any three nodes of the graph is shown in Figure~\ref{fig:example-optimal-strategy}.
	
	\begin{figure}[!ht]
		\centering
		\begin{tikzpicture}
			\node[circle, draw, fill=gray] (A) at (0, 0) {A};
			\node[circle, draw, fill=gray] (B) at (2, 0) {B};
			\node[circle, draw] (C) at (1, 1.7) {C};
			\node[circle, draw] (D) at (1, 3.2) {D};
			\node[circle, draw, fill=gray] (E) at (4, 0) {E};
			\draw (A) -- (B);
			\draw (B) -- (C);
			\draw (C) -- (A);
			\draw (C) -- (D);
			\draw (B) -- (E);
		\end{tikzpicture}
		\caption{3-SRD problem: An attack $P=\{A,B, E\}$ (filled in gray) can be defended by a function $f$ provided with $f(D)=f(A)= f(B)=0$, $f(C)=3$ and $f(E)=1$ with a $\omega(f) =4$. But, the attack $P=\{A,B, D\}$ cannot.} \label{fig:defended-attack}
	\end{figure}
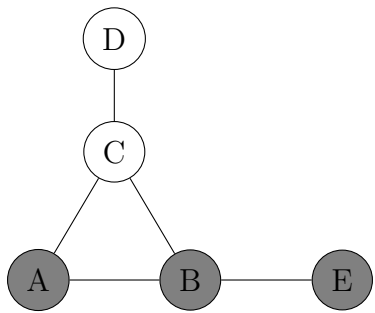

	The assignment of a number $c\in \mathbb{N}_0$ to a node can be seen as determining the number of field armies (FAs) at that node. The aspect of protecting an undefended node -- which has no FAs -- can be interpreted as moving an FA that is at a neighboring location to that (unprotected) location.
	Some direct properties of a valid $k$-SRD function are given as follows. By assigning the value 1 to all nodes in $G$, all nodes \emph{are considered safe for any attack of $\mathcal{P}$}, so that we obtain a trivially valid $k$-SRD function on any graph $G$ and thus an upper bound for the $\gamma_{k-SRD}$-number of a graph of size $n$. All label assignments with only positive values are therefore considered trivially protected. This means that only assignments that contain at least one node labeled with 0 are of interest to us. \\
	
	The rest of the paper is structured as follows. In Section~\ref{sec:vns}, details are given about the designed VNS approach to solve the $k$-SRD problem. In addition, a greedy approach for the $k$-SRD problem is presented. Section~\ref{sec:experimental-evaluation} provides a thorough experimental evaluation between our two heuristic approaches and the two exact literature approaches. Section~\ref{sec:case-study} describes a use case of the $k$-SRD problem using instances extracted from spatial data of GeoJSON files. This paper concludes with section~\ref{sec:conclusion} and an outlook on future work.

	\section{Variable neighborhood search for $k$-SRD problem}\label{sec:vns}
	
	\emph{Variable neighborhood search} is a powerful single-point search meta-heuristic developed by Mladenović and Hansen~\cite{mladenovic1997variable}. It has been applied to numerous difficult computational problems from various fields, such as scheduling~\cite{roshanaei2009variable,abdullah2005investigation}, allocation problems~\cite{hansen2008variable}, routing~\cite{jarboui2013variable,hemmelmayr2009variable}, bioinformatics problems~\cite{charpentier2018variable, grbic2020a}, domination problems~\cite{ivanovic2019variable, kapunac2023} and many others.
	
	The main idea of VNS is to use a set of different neighborhood structures with increased cardinality around a solution to avoid the search getting stuck in local optima. When the algorithm makes no progress, a new (usually increasingly larger) neighborhood structure is placed around the current solution. Various versions of the VNS are known in the literature, such as the variable neighborhood decomposition~\cite{hansen2001variable}, the skewed VNS~\cite{smiti2020skewed}, the general VNS~\cite{mladenovic2008general}, etc. The schematic of the basic VNS is shown in Algorithm~\ref{alg:vns}.
	
	\begin{algorithm}
		\begin{algorithmic}[1]
			\Statex \textbf{Input}: problem instance $I$, $r_{\min}, r_{\max} >0$, a set of neighborhood structures $\mathcal{N}_{r_{\min}}, \ldots, \mathcal{N}_{r_{\max}}$, (fitness) function $f$
			\Statex \textbf{Ouput}: solution $s$
			\State $s \gets \texttt{generate\_initial\_solution}(I)$ 
			\State $s_{best} \gets s$
			\While{!(any of termination criteron is met)}
			\State $r \gets r_{\min}$
			\While{$r \leq r_{\max}$}
			\State $s' \gets \texttt{shake}(s, \mathcal{N}_r)$
			\State $s'' \gets \texttt{local\_search}(s')$
			\State $s_{best}, r \gets \texttt{acceptance\_criterion}(s_{best}, s'', r)$
			\EndWhile
			\EndWhile
			\State \textbf{return} $s_{best}$
		\end{algorithmic}
		\caption{The basic VNS framework}\label{alg:vns}
	\end{algorithm}
	
	The algorithm first initializes a starting solution $s_{best}$, usually randomly, or in a more sophisticated way, e.g. using a greedy procedure or another meta-heuristic that quickly provides a solution of reasonable quality. Subsequently, the main loop is initiated by executing basic VNS iterations as long as one of the termination conditions is not met: allowed time limit, allowed number of VNS iterations exceeded, number of iterations reached without any improvement, etc. A single VNS iteration consists of entering the inner loop in which the neighboring structures are systematically examined, starting with the smallest, i.e. $r \gets r_{\min}$. In each iteration, the operator \texttt{shake}() is executed on the solution $s$ around the $r$-th neighboring structure which returns a modified solution $s'$. This solution is potentially improved by a local search procedure \texttt{local\_search}(). Then, based on the function \texttt{acceptance\_criterion}(), this new solution $s''$ is either accepted or not for a new incumbent, which changes the value of $r$. (Usually, $r \gets r_{\min}$ when $s''$ is accepted.) The inner loop runs until $r$ no longer reaches $r_{\max}$, i.e. when all neighborhood structures around $s_{best}$ have been examined.
	
	The rest of this section is devoted to the details of the design of an efficient VNS for solving the $k$-SRD problem.
	
	\subsection{Predefined attacks and the concept of quasi-feasibility} \label{sec:attacks}

	Before delving into the specifics of the primary procedures that form the core of the main VNS, it is crucial to acknowledge the inherent difficulty in verifying the feasibility of a single solution for this problem in general. This challenge is primarily attributed to the exponentially increasing number of attacks to be verified as the instance size grows. Consequently, we introduce two fundamental concepts designed to address this issue: a set of predefined attacks and the quasi-feasibility. Ultimately, these concepts enable us to tackle large-sized instances from a practical standpoint.

	\subsubsection{Predefined attacks}
	
	We separate sets of attacks to two groups, both related to the concept of quasi-feasibility in general case, although exact feasibility can be reached for the cases of small to medium-sized instances:
	
	\begin{itemize}
		\item \textit{Intensive attacks}, labeled by \emph{intense\_attacks}. These attacks are employed to assess the more confident quasi-feasibility of the current solution.
		\item \textit{Lightweight attacks} labeled by \emph{lightweight\_attacks}. As explained in the subsequent paragraphs attacks of this type serve in the local search procedure to access the concept of relaxed quasi-feasibility.
		
	\end{itemize}
	
	The bound on the number of attacks is parameter \emph{comb\_take\_all\_bound}. In case $\binom{n}{k} <comb\_take\_all\_bound$, we utilize all possible $k$-attacks $\mathcal{P}$ for \emph{intense\_attacks} as well as for \emph{lightweight\_attacks}. Otherwise, we generate a subset $\mathcal{P}' \subseteq \mathcal{P}$ of attacks in the following way. First, for each node $v$, the set $N_l^v $ consists of all nodes which are at the distance of at most $l$ from $v$.  Given a set of nodes $s$ and $k$, by \emph{comb}($s, k$) we simply denote the set of all $k$-combinations of set $s, |s|\geq k$. Further, for fixed node $v$, we generate the set of \emph{intense\_attacks} as the union of $comb(N_3^v, k)$ across each node $v$. We chose $l=3$ for generating these attacks as it provides a good balance between the performance and good approximation of solution feasibility.  In case that the size of \emph{intense\_attacks} does not exceed the value \emph{comb\_take\_all\_bound}, we set \emph{lightweight\_attacks}=\emph{intense\_attacks}, otherwise \emph{lightweight\_attacks} is obtained as the union of $comb(N_1(v), k)$ over all $v$.

	\subsubsection{The concept of quasi-feasibility}\label{sec:quasi}
	
	The concept of \emph{quasi-feasibility} enhances tractability of verifying solution feasibility by focusing on a predefined set of attacks (\emph{intense\_attacks}) instead of verifying exponentially many attacks (all possible attacks). In practical terms, this entails replacing an exponential-time exact feasibility verification with a semi-decidable polynomial verification algorithm. (Semi-decidable in the sense that if the procedure yields a negative answer, the solution is indeed not feasible, and if the procedure yields a positive answer, the feasibility is not guaranteed to be feasible.) Specifically, for each predefined attack in the set \emph{intense\_attacks}, we assess whether the attack can be successfully defended according to the problem definition. If any attack proves to be undefended by our strategy, the solution is considered infeasible; otherwise, it is considered quasi or (weakly) feasible. The essential steps for verifying whether a specific attack can be successfully defended are outlined in Algorithm~\ref{alg:is-defended}.
	
	\begin{algorithm}
		\footnotesize{
			\begin{algorithmic}[1]
				\Statex \textbf{Input}: solution $s$,  $G=(V,E)$, $k$, \emph{attack}, $cutoff>0$, $tries>0$
				\Statex \textbf{Output}: a pair \emph{attack\_defended}, \emph{defending\_nodes}
				
				\State $all\_alternatives, reduced\_attack \gets \emptyset$
				\For{$v \in attack$}
				\If{$s[v] > 0$}
				\State add $v$ to \emph{defending\_nodes}
				\State \textbf{continue}
				\EndIf
				\State $alternatives \gets $ the set of all neighbors of $v$ that can defend it
				\If{$alternatives = \emptyset$}
				\State \textbf{return} $False, \emptyset$
				\EndIf
				\State add $v$ to $reduced\_attack$
				\State $all\_alternatives \gets all\_alternatives \times alternatives$
				\EndFor
				
				\If{$|all\_alternatives| < cutoff$}
				\State \textbf{return} \texttt{determinitic\_defense}($all\_alternatives, reduced\_attack, defending\_nodes$)
				\Else
				\State \textbf{return} \texttt{roulette\_defense}($reduced\_attack, defending\_nodes, tries$)
				\EndIf
		\end{algorithmic}}
		\caption{Procedure \texttt{is\_attack\_defended()}}
		\label{alg:is-defended}
	\end{algorithm}
	
	We have designed two defense strategies aimed at protecting nodes within the specified set of attacks: a deterministic strategy and a probabilistic one, outlined in Algorithms~\ref{alg:determinitic-defense} and~\ref{alg:probabilistic-defense}, respectively, see~\ref{app: pseudocodes} for more details. The choice between these strategies is made based on the number of potential alternative protectors for all undefended nodes and the parameter \emph{cutoff}. If the number of all alternatives is relatively small, we can afford to explore all combinations for defenders. Otherwise, we employ a roulette selection of protector nodes, i.e., nodes that have a greater chance of being selected as protectors for the current node. Since this probabilistic strategy may not succeed every time, we execute it multiple times controlled by the parameter $tries>0$.\\
	
	The core procedure, called \texttt{quasi\_infeasibility}() (Algorithm~\ref{alg:quasi-infeas}), for each attack $h \in intense\_attacks$ executes the function \texttt{is\_attack\_defended}() given in Algorithm~\ref{alg:is-defended}. In case no valid defense is found, the counter that keeps track of the number of non-defended attacks is incremented, which is, at the end, returned by the procedure. 
	
	\begin{algorithm}[H]
		\footnotesize{
			\begin{algorithmic}[1]
				\Statex \textbf{Input}: solution $s$,  $G=(V,E)$, $k$, \emph{attacks}, $cutoff>0$, $tries>0$
				\Statex \textbf{Output}: \emph{non\_defended\_count}, \emph{coverage\_info}
				\State $non\_defended\_count \gets 0$
				\State $coverage\_info \gets \emptyset$
				\For{$attack \in attacks$}
				\State $defended, defending\_nodes \gets$ is\_attack\_defended($s$,  $G$, $k$, $attack$, $cutoff$, $tries$)
				\If{$defended$}
				\State \texttt{update\_coverage\_info}($defending\_nodes$, $attack$)
				\Else
				\State $non\_defended\_count \gets  non\_defended\_count+1$
				\EndIf
				\EndFor
		\end{algorithmic}}
		\caption{Procedure \texttt{quasi\_infeasibility()}}
		\label{alg:quasi-infeas}
	\end{algorithm}
	
	\subsection{Initialization: a greedy approach} \label{sec:greedy}
	
	To initialize a feasible solution, we have devised an efficient greedy approach based on the principle of covering nodes of a graph. In the following section, we provide a detailed description of our greedy approach, pseudo-coded in Algorithm~\ref{alg:greedy-cover}.

	Initially, all nodes are assigned a label 0, represented as the solution $s = [0, \ldots, 0]$ and the coverage $\emph{covered}=\emptyset$ which keeps nodes that are already protected under any attack. In each iteration, the algorithm decides which node to label based on a greedy criterion $g(\cdot)$. This criterion calculates, for each not-yet-processed node, the number of its neighboring nodes (counting also itself) still not included in the set \emph{covered}. Among such nodes, the algorithm selects one with the highest $g$-value, denoted as $v'$. Symbolically, the greedy value for a node $v \in V(G) \setminus \emph{covered}$ is determined by:
	$$ g(v) = | N[v] \setminus \emph{covered}|$$
	
	If there are multiple nodes with equal $g$-value, we prioritize one not belonging to \emph{covered}. The reasoning behind choosing this type of tie breaking is based on the following facts.
	\begin{itemize}
		\item If a node $v$ has, for example, $g(v) = 3$ and $v$ itself is not covered, then it has to cover itself and two of its neighbors, so a total of $3$ FAs is sufficient.
		\item On the other hand, if the node $v$ is covered and has $g(v) = 3$, that means it has three neighbors to cover and, as always, needs one FA for itself, resulting in $4$ FAs in total.
	\end{itemize}
	
	Then, a label $l_{v'}=\min(k + 1, g(v') + I_{v' \in covered})$ is assigned to $v'$, that is $s[v']=l_{v'}$, where $I$ is an indicator function. Consequently, all neighbors of $v'$ immediately become covered -- that is included in the set \emph{covered}. The algorithm terminates when all nodes are covered.

	\begin{algorithm}
		
		\begin{algorithmic}[1]
			\Statex \textbf{Input}: an instance $I$ of $k$-SRD problem, k
			\Statex \textbf{Output}: solution $s$
			\State $ s \gets [0, \ldots, 0]$
			\State $covered \gets \emptyset$
			\While{$ covered \neq V(G)$}
			\State $v' \gets \arg \max \{ g(v) \mid v \in V(G) \setminus covered \}$ \Comment{tie breaking - uncovered nodes go first}
			\State $uncov_{v'} \gets N[v'] \setminus covered $
			\If{$v' \in uncov_{v'}$}
			\State $v'\_not\_counted \gets 0$
			\Else
			\State $v'\_not\_counted \gets 1$
			\EndIf
			\State $s[v'] \gets \min(k + 1, g(v') + v'\_not\_counted)$
			\State $covered \gets covered \cup uncov_{v}$
			~
			\EndWhile
		\end{algorithmic}
		\caption{Procedure \texttt{greedy()}}
		\label{alg:greedy-cover}
	\end{algorithm}
	
	\subsection{Shaking}\label{sec:shake}

	This phase is in charge of diversifying the search process of the VNS algorithm. It is realized through the procedure $\texttt{shake}(s, r)$, where the input parameters consist of a solution $s$ and the $r$-th neighborhood structure $\mathcal{N}_r$. The output of this procedure yields a (mutated) solution $s'$ generated in the following way. For a given solution $s$ that is to be \emph{shaked}, we perform $r$ random increments and $r+1$ random decrements where decrement is performed on non-zero labels only. The choice of decrements is based on the roulette selection  of the labels of solution $s$.  
	For example, given a solution $s=[2,3,4,1]$ and $r=2$, after two random increments solution might be $s'=[3, 3, 4, 2]$ and after 3 decrements, it might be $s'=[1, 3, 3, 2]$. Therefore, $\omega(s)-\omega(s')=1$.
	
	\subsection{Local search (LS)}\label{sec:ls}
	
	The primary objective of local search is to intensify the exploration of solutions in the vicinity of a given solution $s$ with the aim of finding high-quality local optima. For the $k$-SRD problem, we have devised a suitable local search procedure guided by two key principles: ($i$) the \textit{quasi-swap principle}; and ($ii$) the \textit{first-improvement} strategy. As the LS can take considerable amount of time, feasibility of solutions generated through the LS process is verified in a relaxed fashion by using the set \emph{lightweight\_attacks} of predefined attacks.
	
	The quasi-swap principle involves first shuffling the set of node pairs $(i, j)$, where $i < j$, before proceeding to the next step to eliminate any potential biases. Subsequently, for each node pair $(i, j)$, we apply partial-imputations of FAs from the set of 2-decomposition of the sum $s[i] + s[j]$. For instance, if $s[1]=2$, and $s[3]=3$, we construct the set of 2-decomposition for $2+3=5$: ${(0, 5), (1, 4), (3, 2), (4, 1), (5, 0)}$. Then, for each 2-decomposition, replacing $s[i]$ and $s[j]$ with a respective component value is probed to generate a new solution candidates $\{(s[1]=0,s[3]=5), (s[1]=1, s[3]=4), ...\}$. Afterwards, quasi-feasibility of a solution candidate (see Section~\ref{sec:quasi}) is verified, returning number of non-defended attacks (under the set of reduced, predefined attacks leveraged for an incremental fitness evaluation, see the next section). The solution is accepted as a new incumbent if the number of non-defended (reduced) attacks decreases. If no improvement is observed in that way, next 2-decomposition is probed to replace the current best solution's components, and the process is repeated. If none of the 2-decomposition replacement yields an improvement over the incumbent solution, a different node pair $(i, j)$ is chosen, and the process is restarted. 
	
	If no improvement occurs with any node pair, the LS terminates unsuccessfully returning the same input solution $s$ along with its fitness score. Otherwise, if the number of non-defended attacks reaches 0 during the search, indicating the achievement of a (quasi-)feasible solution, the algorithm terminates successfully, returning the improved solution along with its new improved fitness score. 
	
	\subsection{Fitness function}\label{sec:fitness}
	
	The fitness function $fitness$ for a $k$-SRD solution $s$ comprises two scores in the form of an ordered pair $fitness(s) = (\emph{quasi-infeasibility}(s), \emph{sum}(s))$, where:
	
	\begin{itemize}
		\item $\emph{quasi-infeasibility}(s)$ denotes the number of non-defended attacks for solution $s$, and
		\item $\emph{sum}(s)$ represents the sum of all elements in $s$.
	\end{itemize}
	
	For the LS, we have implemented a fast fitness score calculation method structured as follows. During each execution of \texttt{quasi\_infeasibility}(), the coverage information is determined. This entails storing information within an appropriate data structure, referred to as \emph{coverage\_info}, when node $v$ is utilized as a defending node for an attack $h \in \textit{lightweight\_attacks}$.  Leveraging this structure has proven to significantly accelerate the LS process, as it allows for rapid verification of some attacks in \textit{lightweight\_attacks} as protected for solutions $s'$ in the vicinity of solution $s$. 

	\subsection{The core VNS scheme for the $k$-SRD problem}
	
	We finally conclude the overall outline of the proposed VNS for $k$-SRD in Algorithm~\ref{alg:vns-k-srdp}.
	
	\begin{algorithm}[H]
		
		\footnotesize{
			\begin{algorithmic}[1]
				\Statex \textbf{Input}:  an instance $I$ of  $k$-SRD problem, $k, r_{\min}, r_{\max}, move_{prob}, t_{\max}, iter_{max}, tries, cutoff$
				\Statex \textbf{Output}:  (possibly feasible) solution $s_{best}$
				\State $intense\_attacks, lightweight\_attacks \gets \texttt{generate\_attacks}(I, k)$ \Comment{see Section~\ref{sec:attacks}}
				\State $s_{best} \gets \texttt{greedy}(I, k)$ \Comment{see Section~\ref{sec:greedy}}
				\State $iter \gets 0$
				\State $infeasibility_{s_{best}} \gets \texttt{quasi\_infeasibility}(s_{best},  intense\_attacks, cutoff, tries)$ \Comment{see Section~\ref{sec:quasi}}
				\State $fitness_{s_{best}} \gets (infeasibility_{s_{best}}, sum(s_{best}))$
				\While{$t_{passed} < t_{\max}$ and $iter < iter_{max}$}
				\For{$r=r_{\min} \text{ to } r_{\max}$}
				\State $iter \gets iter+1$
				\State $s' \gets \texttt{Shake}(s_{best}, r)$ \Comment{see Section~\ref{sec:shake}}
				\State $s', fitness_{s'} \gets \texttt{LS}(s', lightweight\_attacks, tries)$ \Comment{see Section~\ref{sec:ls}}
				\If{$fitness_{s'} < fitness_{s_{best}}$ or ($fitness_{s'} = fitness_{s_{best}}$
					and $move_{prob} < u \in \mathcal{U}(0, 1)$)}
				\State $intense\_infeasibility_{s'} \gets \texttt{quasi\_infeasibility}(s', intense\_attacks, cutoff, tries) $
				\If{$intense\_infeasibility_{s' } = 0$} \Comment{solution is quasi-feasible}
				\State $s_{best} \gets s'$
				\State $fitness_{s_{best}} \gets (intense\_infeasibility_{s'}, sum(s')) $
				\State \textbf{break}
				\EndIf
				\EndIf
				\EndFor
				\EndWhile
		\end{algorithmic}}
		\caption{VNS algorithm for the $k$-SRD problem}\label{alg:vns-k-srdp}
	\end{algorithm}
	
	In the first line of Algorithm \ref{alg:vns-k-srdp}, the tasks outlined in Section~\ref{sec:attacks} are executed. Specifically, two distinct types of attacks, denoted as $intense\_attacks$ and $lightweight\_attacks$, are generated. Subsequently, the greedy algorithm from Section~\ref{sec:greedy} is employed to produce an initial solution $s_{best}$. This solution is then subjected to verification via $\texttt{quasi\_infeasibility}()$, as outlined in Algorithm~\ref{alg:is-defended}. The resulting count of non-defended attacks is stored in $\textit{infeasibility}_{s_{best}}$ and its fitness score in $fitness_{s_{best}}$.

	The main loop of the VNS algorithm is initiated, and the algorithm continues its execution until either the time limit is reached or the maximum allowed number of iterations is exceeded. The basic iteration of VNS comprises the following steps. Initially, the set of neighborhood structures is systematically explored, commencing with the smallest one, denoted as $r_{\min}$. Within the current neighborhood structure $r$, the \texttt{Shake}() procedure (refer to Section~\ref{sec:shake}) is executed, generating solution $s'$ which is subsequently refined by the local search  \texttt{LS}() (see Section~\ref{sec:ls}). Next, the current incumbent solution $s_{best}$ is compared with $s'$ (see Lines 11-18). If the number of non-protected attacks in $s'$ is reduced compared to $s_{best}$, or if they match but the sum of all labels in $s'$ is strictly smaller, then solution $s'$ undergoes further verification. This verification involves executing the $\texttt{quasi\_infeasibility}()$ procedure with the set \emph{intense\_attacks}. Furthermore, if the fitness values of $s'$ and $s_{best}$ are equal, the verification of quasi-feasibility for $s'$ is applied with a probability controlled by the parameter $move_{prob}$. Once the (quasi-)feasibility of $s'$ has been verified, if the number of non-protected attacks is reduced to 0, $s'$ is accepted as the new incumbent solution. Subsequently, the next iteration of VNS proceeds with the smallest neighborhood structure. If all neighborhood structures are exhausted ($r$ exceeds $r_{\max}$), indicating the completion of the primary VNS iteration, the subsequent iteration starts again from the smallest neighborhood structure, that is $r_{\min}$.
	
	\section{Experimental evaluation}\label{sec:experimental-evaluation}
	
	In this section we present a comprehensive experimental evaluation comparing our VNS with the competitor approaches from the literature, specifically with ILP and the Benders-based decomposition approach described in \cite{liu2020k}; the latter approach is denoted simply as \textsc{Benders}. Additionally, we include the greedy method from Section~\ref{sec:greedy}, labeled as \textsc{Greedy}, to assess the improvement factor of  solutions obtained by VNS over its initial solutions produced via the greedy approach.
	The source codes of the competitor approaches (ILP and \textsc{Benders}) were obtained from the first author of the paper \cite{liu2020k}. 
	VNS and \textsc{Greedy}  approaches are implemented in $C$++ and executed on Ubuntu 20 while compiled with $C$++17 compiler involving the $Ofast$ optimization level. To facilitate program execution, we have enabled running the algorithm from a program written in Python (version 3.9), which imports appropriate nested modules linking and implicitly executing necessary parts of the original $C$++ program. In that way, we supported  installation of our algorithms by simply running the following command: \texttt{pip install ksrdp}.  Moreover, our source code along with the instances and raw results are freely available and can be downloaded from the GitHub repository at the link \url{https://github.com/StefanKapunac/ksrdp}.
	
	All our experiments were performed in single-threaded mode on an Intel Xeon E5-2640 with 2.40GHz and 16 GB of memory per each execution.
	
	\subsection{Problem instances}
	
	Three benchmark sets were used for our experimental evaluation: \textsc{Random}, \textsc{Wireless}, and \textsc{Real} sets. The benchmark set \textsc{Random} comprises randomly generated graphs initially proposed in~\cite{liu2020k}. For each graph size $n \in  \{10, 15, 20, 30, 45,  50, 100\}$ there are 5 graphs of different densities which, in overall, gives, 35 problem instances. For more information about the properties of these graphs, we refer interested readers to the aforementioned paper.  In case of the  benchmark set \textsc{Wireless}, 16 instances are created, corresponding to ad-hoc wireless networks of various densities. To generate these graphs, we utilized the Unit disc model~\cite{clark1990unit}. This model generates random geometric graphs based on two parameters $n$ and $R\in (0, 1)$, representing the graph size and the radius value, respectively. Specifically, the algorithm first randomly places $n$ Euclidean points within the unit square in the plane, which serve as the nodes of the graph. Subsequently, edges are created to connect all pairs of points (nodes) that are within a distance of at most $R$. We generated a total of 16 instances, with $n \in \{20, 30, 50, 100\}$ and $R \in \{0.3, 0.4, 0.5, 0.6\}$. For generating these ad-hoc wireless networks, we used the known \texttt{NetworkX} module in Python, specifically leveraging the \texttt{random\_geometric\_graph()} function.
	
	\begin{figure}[H]
		\centering
		\includegraphics[width=280pt, height=200pt]{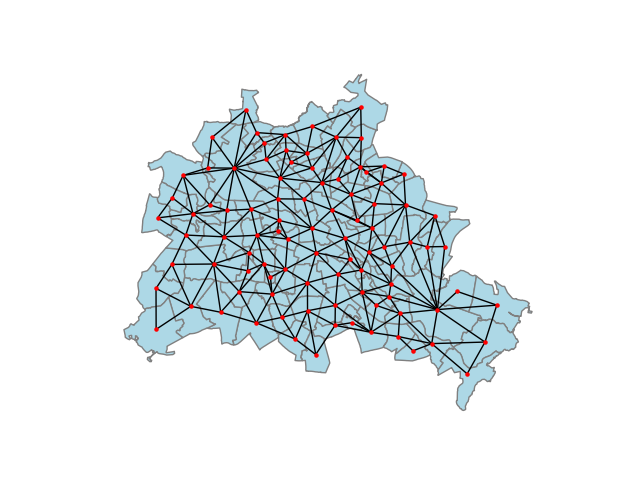}
		\caption{The city of Berlin divided into districts: the \emph{Queen} adjacency graph }
		\label{fig:queen-city}
	\end{figure}

	Regarding the benchmark set \textsc{Real}, the instances are specifically designed to support the case study surrounding the real-world application of the $k$-SRD problem as given in Section~\ref{sec:case-study}. To generate these graphs, we utilized a collection of GeoJSON files originated from  the GitHub repository~\url{https://github.com/blackmad/neighborhoods.git}. These (168) files encode geographical data structures used to represent geographic features such as points, lines, and polygons. In our context, they represent cities divided into districts, countries divided into regions, or even continents divided into countries, among other possibilities. To generate graphs corresponding to these files, we utilized two Python modules, PySAL and GeoPandas, explicitly developed to work with GeoJSON data. The process of generating graph instances from these files works as follows. First, for each polygon in a file, its centroid point is determined, representing a node of the corresponding graph. Subsequently, the \emph{Queen} adjacency graph is constructed~\cite{griffith2011approximating}, where   two centroids are linked if the two polygons corresponding to these centroids share at least one point on their boundary, meaning their intersection is non-empty.
	
	In Figure~\ref{fig:queen-city}, one can see an illustration of a graph instance  generated for a specific GeoJSON file containing information about the division of the city of Berlin into districts. 
	
	\subsection{The parameter tuning of the VNS approach}
	
	Based on preliminary results, we categorized the instances into three groups: ($i$) \emph{small-sized}, consisting of instances with $n < 30$; ($ii$) \emph{middle-sized}, comprising instances with $30 \leq n < 100$; and ($iii$) \emph{large-sized}, otherwise.
	
	For \emph{small-sized} instances, a time limit of 300 CPU seconds was set; for \emph{middle-sized} instances, the time limit was extended to 600 CPU seconds, and for \emph{large-sized} instances, it was further increased to 1200 CPU seconds, i.e. 20 minutes.  It is important to note that the same time limit was applied to all four approaches. In addition, the parameter $num\_iter$ (the iteration limit criterion) of the VNS approach was set to 5000 for all instances.
	
	The rest of parameter values utilized in our VNS across all instances were: $r_{\min}=1$, $r_{\max}=10$, $move\_{prob}=0.5$, $cutoff=100$, and $tries=10$.

	\subsection{Experimental results}
	
	This section reports the experimental results of the four competitors. We consider $k \in \{2, 3, 4, 5\}$ for all our experiments following the same  experimental methodology from~\cite{liu2020k}.
	The results on benchmark set \textsc{Random} are presented in seven tables, grouped according to different value of $n \in \{10,15,20, 30, 45, 50, 100\}$. The results on benchmark set \textsc{Wireless} are reported in four tables, one table per each  value of $k$.
	
	\subsubsection{Results on benchmark set \textsc{Random}}
	
	Experimental results on benchmark set \textsc{Random} for $n\in \{30, 50, 100\}$ are shown in Tables~\ref{tab:results-random-n-30}--\ref{tab:results-random-n-100}. The remaining results are given in \ref{app: random-remaining-results}, see Tables~\ref{tab:results-random-n-10}--\ref{tab:results-random-n-45}. These tables are organized as follows.  The first two columns provide characteristics of respective instance in terms of the graph size ($n$) and its density (that refers to column \emph{index}), for the details of the generation and how these instances are labeled, we refer to~\cite{liu2020k}.  The third column represents the value of $k$.  Next four blocks report the results of \textsc{Greedy}, \textsc{Vns}, \textsc{Ilp} and \textsc{Benders} approach, respectively. For each of two exact approaches and the related blocks, two statistics are shown:
	the obtained solution quality  (column ${obj}$) and the overall running times (column $t[s]$).  For \textsc{Greedy} approach, its running times are omitted as these were consistently less than 1 second, thus significantly lower than the other three approaches. For the \textsc{Vns} approach, apart from the reported average solution quality and average running times of achieving best solutions over 10 runs for each instance, additional information is shown -- (relative) standard deviation (in percentage) for solution quality over 10 runs (column $\sigma[\%]$). In case of  \textsc{Benders} approach,  the obtained dual bound (column \emph{dual}) is additionally provided.  The best obtained results among the four competitor approaches are displayed in bold. \\
	
	The following conclusions may be drawn from the obtained numerical results reported in Tables~\ref{tab:results-random-n-30}--\ref{tab:results-random-n-100}.
	
	\begin{itemize}
		\item For the instances with $n=30$ and small $k\in \{2,3\}$ one could notice that \textsc{Ilp} was able to find a proven optimal solution for all problem instances while its running times increase dramatically when moving from $k=2$ to $k=3$. The \textsc{Benders} approach was running out of the time limit for two cases but delivering  dual (lower) bounds which  match the value of the corresponding optimal solutions. Concerning our \textsc{Vns} approach, it delivers solutions whose quality is equal to the known optimal solutions. In most cases, its running times are comparable to those of the \textsc{Ilp} approach. Moreover, the initial solutions of \textsc{Vns} obtained by \textsc{Greedy} approach are significantly improved in the search phase of \textsc{Vns}.
		\item For the instances with $n=30$ and larger $k\in \{4,5\}$, \textsc{Ilp} could not even construct corresponding ILP model within the given memory limit. The \textsc{Benders} approach could not find a proven optimal solution for any of the problem instances. Average solutions  quality of \textsc{Vns} approach is significantly better than the quality of solutions provided by the \textsc{Greedy} approach. One can notice that the lower bounds reported by the Benders approach lead to the conclusion that the solutions of VNS are not far from the optimal ones, if not matching the optimal ones.
		\item For the instances with $n=50$ and  $k=2$ one could notice that \textsc{Ilp} could find a proven optimal solution for all 5 problem instances. This was possible for \textsc{Benders} approach on 3 problem instances whereas for two remaining cases it is running out of the given time limit.  The average solution quality of \textsc{Vns} approach  matches the quality of optimal solutions in 4 (out of 5) problem instances and failing to do it for just one case (29.3 versus 29). In all cases, \textsc{Vns} was able to significantly improve over the initial solutions obtained by \textsc{Greedy} approach.
		\item For the instances with $n=50$ and  $k \in \{3, 4, 5\}$ one could see that the exact   \textsc{Ilp} and \textsc{Benders} approaches could not find a proven optimal solution for any problem instances due to occurred memory issues of initializing ILP model and reaching the given time limit to run the algorithm, respectively. Our \textsc{Vns} approach again has delivered significantly better (average) results than those of \textsc{Greedy} approach. 
		\item For the instances with $n=100$ and  $k=2$ we notice that \textsc{Ilp} could find a proven optimal solution for all (5) problem instances whereas \textsc{Benders} approach could find none of them along with delivering relatively weak dual bounds. The (average) quality of solutions obtained by \textsc{Vns} are within $\approx$ 4\% gap from the quality of known optimal solutions. Moreover, \textsc{Vns} significantly improves over the initial solutions provided by \textsc{Greedy} approach.
		
		\item For the instances with $n=100$ and  $k \in \{3, 4, 5\}$, \textsc{Ilp} cannot find a proven optimal solution due to memory issues. The \textsc{Benders} approach is running out of time and delivering consistently  weak dual bounds. Again, the obtained (averaged) solutions of \textsc{Vns} are significantly better than those of \textsc{Greedy} approach -- in some case the relative improvements are around 10\%.
	\end{itemize}
	\begin{table}
	\centering
	\footnotesize{
		\begin{tabular}{rrrrrrrlllll}
			\hline \hline
			\multicolumn{2}{c}{Instance} &
			\multicolumn{1}{c}{$k$} &
			\multicolumn{1}{c}{\textsc{Greedy}}   &
			\multicolumn{3}{c}{\textsc{VNS}}      &
			\multicolumn{2}{c}{\textsc{Ilp}}      &
			\multicolumn{3}{c}{\textsc{Benders}} \\  
			\cmidrule(rr){1-2}   \cmidrule(r){4-4} \cmidrule(rrr){5-7} \cmidrule(rr){8-9} \cmidrule(rrr){10-12}

			$n$ & \textit{index} & & obj & $\overline{obj}$ &  $\sigma[\%]$ & $\overline{t}_{best}[s]$ & obj & $t[s]$ & obj & dual &  $t[s]$ \\ \hline
30 & 1 & 2 & 20 & \textbf{17.0} 	&0.00 & 19.0 & \textbf{17} & 4.7 & \textbf{17} & 17 & 41.2 \\
 & 2 &  & 20 & \textbf{17.0} 		&0.00 & 1.6 & \textbf{17} & 6.5 & \textbf{17} & 17 & 28.4 \\
 & 3 &  & 21 & \textbf{17.0} 		& 2.8& 10.6 & \textbf{17} & 8.3 & \textbf{17} & 17 & 87.4 \\
 & 4 &  & 17 & \textbf{16.0}		& 0.0& 0.3 & \textbf{16} & 1.1 & \textbf{16} & 16 & 36.0 \\
 & 5 &  & 18 & \textbf{17.0}		& 0.0& 4.5 & \textbf{17} & 6.6 &  \textbf{17} & 17 & 23.3 \\ \hline
30 & 1 & 3 & 23 & \textbf{19.0}		& 0.0& 249.4 & \textbf{19} & 208.9 & 0 & 19 & TL \\
& 2 &  & 23 & \textbf{20.0}			& 0.0 & 45.1 & \textbf{20} & 222.2 & \textbf{20} & 20 & 373.6 \\
 & 3 &  & 25 & \textbf{20.0}		& 0.0& 3.7 & \textbf{20} & 380.5 & 0 & 19 & TL \\
 & 4 &  & 21 & \textbf{19.0}		& 0.0& 4.7 & \textbf{19} & 185.2 & \textbf{19} & 19 & 423.9 \\
 & 5 &  & 22 & \textbf{20.0} 		& 0.0& 29.6 & \textbf{20} & 218.1 & \textbf{20} & 20 & 491.9 \\ \hline
30 & 1 & 4 & 26 & \textbf{20.8}		& 2.0& 126.1 & - & - & 0 & 19 & TL \\
 & 2 &  & 26 & \textbf{21.0} 		& 0.0& 104.8 & - & - & 0 & 20 & TL \\
 & 3 &  & 29 & \textbf{21.0}		& 0.0& 374.3 & - & - & 0 & 19 & TL \\
 & 4 &  & 24 & \textbf{20.1} 		& 1.6& 40.7 & - & - & 0 & 19 & TL \\
 & 5 &  & 25 & \textbf{21.3} 		& 2.2& 554.0 & - & - & 0 & 20 & TL \\ \hline
30 & 1 & 5 & 28 & \textbf{22.4} 	& 2.3& 346.9 & - & - & 0 & 19 & TL \\
 & 2 &  & 29 & \textbf{23.5}		& 4.1& 47.6 & - & - & 0 & 20 & TL \\
 & 3 &  & 30 & \textbf{23.2}		& 2.7 & 452.8 & - & - & 0 & 19 & TL \\
 & 4 &  & 26 & \textbf{21.6}		& 3.2 & 202.2 & - & - & 0 & 19 & TL \\
 & 5 &  & 27 & \textbf{23.0}		& 2.0 & 63.6 & - & - & 0 & 20 & TL \\ \hline \hline
\end{tabular}}
\caption{Comparison of the approaches on benchmark set \textsc{Random} for $n=30$. } \label{tab:results-random-n-30}
\end{table}

	\begin{table}
	\centering
	\footnotesize{
		\begin{tabular}{rrrrrrrlllll}
			\hline \hline
			\multicolumn{2}{c}{Instance} &
			\multicolumn{1}{c}{$k$} &
			\multicolumn{1}{c}{\textsc{Greedy}}   &
			\multicolumn{3}{c}{\textsc{VNS}}      &
			\multicolumn{2}{c}{\textsc{Ilp}}      &
			\multicolumn{3}{c}{\textsc{Benders}} \\
			\cmidrule(rr){1-2}   \cmidrule(r){4-4} \cmidrule(rrr){5-7} \cmidrule(rr){8-9} \cmidrule(rrr){10-12}	 		
			
			$n$ & \textit{index} & & obj & $\overline{obj}$ &  $\sigma[\%]$ & $\overline{t}_{best}[s]$ & obj & $t[s]$ & obj & dual &  $t[s]$ \\ \hline
50 & 1 & 2 & 31 & \textbf{29.0}		&0.0  & 25.5 & \textbf{29} & 32.9 & 0 & 28 & TL \\
 & 2 &  & 31 & \textbf{28.0}		& 0.0 & 13.1 & \textbf{28} & 15.2& \textbf{28} & 28 & 439.6\\
 & 3 &  & 34 & \textbf{28.0}		& 0.0 & 60.7 & \textbf{28} & 28.7 & \textbf{28} & 28 & 399.8 \\
 & 4 &  & 33 & \textbf{28.0}		& 0.0& 15.6 & \textbf{28} & 21.6 & \textbf{28} & 28 & 375.8  \\
 & 5 &  & 36 & {29.3}				& 0.0 & 120.6 & \textbf{29} & 25.2 & 0 & 28 & TL \\ \hline
50 & 1 & 3 & 36 & \textbf{33.6}		& 1.6 & 525.6 & - & - & 0 & 28 & TL \\
 & 2 &  & 36 & \textbf{33.0}		& 1.5& 339.8 & - & - & 0 & 28 & TL \\
 & 3 &  & 40 & \textbf{32.4}		& 1.4& 246.5 & - & - & 0 & 28 & TL \\
 & 4 &  & 38 & \textbf{32.5}		& 2.2& 157.4 & - & - & 0 & 28 & TL \\
 & 5 &  & 43 & \textbf{34.3}		& 3.0& 304.4 & - & - & 0 & 29 & TL \\ \hline
50 & 1 & 4 & 40 & \textbf{35.6}		& 2.0& 312.9 & - & - & 0 & 28 & TL \\
 & 2 &  & 40 & \textbf{35.6}		& 2.0& 34.1 & - & - & 0 & 28 & TL \\
 & 3 &  & 44 & \textbf{36.9}		& 2.4& 91.5 & - & - & 0 & 28 & TL \\
 & 4 &  & 42 & \textbf{35.1}		& 1.5& 18.2 & - & - & 0 & 28 & TL \\
 & 5 &  & 47 & \textbf{38.1}		& 2.5& 564.7 & - & - & 0 & 28 & TL \\ \hline
50 & 1 & 5 & 44 & \textbf{40.6}		& 2.9& 174.9 & - & - & 0 & 28 & TL \\
 & 2 &  & 44 & \textbf{40.0} 		& 2.1& 92.3 & - & - & 0 & 28 & TL \\
 & 3 &  & 48 & \textbf{42.3} 		& 2.4& 262.7 & - & - & 0 & 27 & TL \\
 & 4 &  & 45 & \textbf{40.3} 		& 2.5& 130.3 & - & - & 0 & 28 & TL \\
 & 5 &  & 50 & \textbf{42.9} 		& 2.9& 406.3 & - & - & 0 & 29 & TL \\ \hline \hline
\end{tabular}}
\caption{Comparison of the approaches on benchmark set \textsc{Random} for $n=50$. } \label{tab:results-random-n-50}
\end{table}

	\begin{table}
	\centering
	\footnotesize{
		\begin{tabular}{rrrrrrrlllll}
			\hline \hline
			\multicolumn{2}{c}{Instance} &
			\multicolumn{1}{c}{$k$} &
			\multicolumn{1}{c}{\textsc{Greedy}}   &
			\multicolumn{3}{c}{\textsc{VNS}}      &
			\multicolumn{2}{c}{\textsc{Ilp}}      &
			\multicolumn{3}{c}{\textsc{Benders}} \\ 
			\cmidrule(rr){1-2}   \cmidrule(r){4-4} \cmidrule(rrr){5-7} \cmidrule(rr){8-9} \cmidrule(rrr){10-12}	 
			$n$ & index & & obj & $\overline{obj}$ &  $\sigma[\%]$ & $\overline{t}_{best}[s]$ & obj & $t[s]$ & obj & dual &  $t[s]$ \\ \hline
100 & 1 & 2 & 65 & 58.3		& 1.2 & 420.4 & \textbf{57} & 402.6 & 0 & 49 & TL \\
 & 2 &  & 68 &     60.4		& 1.2 & 586.8 & \textbf{58} & 752.2  & 0 & 49 & TL \\
 & 3 &  & 66 &     62.1		& 0.9& 499.8 & \textbf{60} & 775.1 & 0 & 52 & TL \\
 & 4 &  & 64 &     60.4		& 2.1& 445.0 & \textbf{58} & 296.5 & 0 & 50 & TL \\
 & 5 &  & 61 &     58.9		& 0.5& 188.0 & \textbf{57} & 315.7 & 0 & 50 & TL \\ \hline
100 & 1 & 3 & 79 & \textbf{74.5}		& 2.8& 696.4 & - & - & 0 & 49 & TL \\
 & 2 &  & 82     & \textbf{78.2}		& 2.4& 782.6 & - & - & 0 & 49 & TL \\
 & 3 &  & 79     & \textbf{76.8}		& 1.3& 715.6 & - & - & 0 & 51 & TL \\
 & 4 &  & 77     & \textbf{75.9}		& 1.1& 511.0 & - & - & 0 & 50 & TL \\
 & 5 &  & 72     & \textbf{69.5}		& 1.0& 483.2 & - & - & 0 & 49 & TL \\ \hline
100 & 1 & 4 & 86 & \textbf{77.7}		& 3.0& 810.8 & - & - & 0 & 49 & TL \\
 & 2 &  & 92     & \textbf{82.8}		& 1.4& 831.1 & - & - & 0 & 49 & TL \\
 & 3 &  & 89     & \textbf{82.6}		& 1.8& 702.7 & - & - & 0 & 51 & TL \\
 & 4 &  & 86     & \textbf{79.1}		& 2.2& 690.1 & - & - & 0 & 50 & TL \\
 & 5 &  & 81     & \textbf{75.8}		& 2.0& 684.8 & - & - & 0 & 50 & TL \\ \hline
100 & 1 & 5 & 92 & \textbf{89.0}		& 2.2& 740.8 & - & - & 0 & 49 & TL \\
 & 2 &  & 98     & \textbf{90.8}		& 1.7& 1035.9 & - & - & 0 & 49 & TL \\
 & 3 &  & 95     & \textbf{92.5}		& 1.5& 804.8 & - & - & 0 & 51 & TL \\
 & 4 &  & 94     & \textbf{91.4}		& 1.6& 840.9 & - & - & 0 & 49 & TL \\
 & 5 &  & 88     & \textbf{85.5}		& 1.4& 815.2 & - & - & 0 & 49 & TL \\ \hline \hline
\end{tabular}}
\caption{Comparison of the approaches on benchmark set \textsc{Random} for $n=100$. } \label{tab:results-random-n-100}
\end{table}

	\subsubsection{Results on benchmark set \textsc{Wireless}}
	
	Experimental results on this benchmark set are grouped by different values of $k$ and thus provided in four different tables (one table per each value of $k$). The results for $k=3$ and $k=4$ are displayed in Tables~\ref{tab:results-wireless-k-3} and \ref{tab:results-wireless-k-4} respectively; the rest of the results can be found in Tables~\ref{tab:results-wireless-k-2}--\ref{tab:results-wireless-k-5} of \ref{app: wireless-remaining-results}. The tables are organized similarly as those from the previous section except the columns describing the characteristics of an instance. Namely, the first column provides the size of a graph ($n$) and the second value of radius ($R$) indicates the density of a graph (the larger the radius, the larger the density of a graph is). The best obtained results among the four competitor approaches are displayed in bold. \\
	
	The following conclusions may be drawn from these results.
	
	\begin{itemize}
		\item For the instances with $n=20$ and $k=3$, one can see that \textsc{Ilp} and \textsc{Benders} approaches are efficient in solving them all to proven optimality. Our \textsc{Vns} approach could find an optimal solution with a remarkably short running time. The results of \textsc{Greedy} approach seem not to deviate much from the optimal results.
		Note that the instances with a larger density are easier to be solved.
		\item  For the larger instances with $n \geq 50$ and $k=3$, the exact approaches could not find a proven optimal solution, except for the one problem instance where the \textsc{Benders} approach succeeded. This is due to the encountered issues with reaching the memory or time limit.  In seven (out of eight) cases the obtained solutions of \textsc{Vns} are significantly better than that of \textsc{Greedy} approach.
		\item For the smallest instances with $n=20$ and $k=4$, the exact \textsc{Ilp} approach struggles to solve these instances achieving a success in case of one problem instance ($n=20$ and $R=0.3$).
		\textsc{Benders} approach demonstrates better performance by solving three (out of four) problem instances. Our \textsc{Vns} finds optimal solutions for all of these problem instances within the time of 1 minute.
		\item   For the larger instances with $n\geq 50$ and $k=4$,  the exact \textsc{Ilp} and \textsc{Benders} approaches could not solve  any of the problem instances due to exceeding the provided memory and  time resources, respectively. Our \textsc{Vns} could find a better solution on six (out of eight) problem instances when compared to the solutions obtained by \textsc{Greedy} approach.
	\end{itemize}
	
	\begin{table}
	 \centering
	\footnotesize{
		\begin{tabular}{rrrrrrlllll}
			\hline \hline
			\multicolumn{2}{c}{Instance} &
			\multicolumn{1}{c}{\textsc{Greedy}}   &
			\multicolumn{3}{c}{\textsc{VNS}}      &
			\multicolumn{2}{c}{\textsc{Ilp}}      &
			\multicolumn{3}{c}{\textsc{Benders}} \\
			
			\cmidrule(rr){1-2}   \cmidrule(r){3-3} \cmidrule(rrr){4-6} \cmidrule(rr){7-8} \cmidrule(rrr){9-11}	 
			
			$n$ & $R$ &  obj & $\overline{obj}$ & $\sigma[\%]$ & $\overline{t}_{best}[s]$ &  obj & $t[s]$  & obj & dual & $t[s]$ \\ \hline
20 & 0.3 &  15 & \textbf{14.0}         &  0.0 & 0.1 & \textbf{14} & 5.73 & \textbf{14} & 14 & 31.1  \\
 & 0.4 &  \textbf{12} & \textbf{12.0}	& 0.0& 0.0 & \textbf{12} & 51.11 & \textbf{12} & 12 & 101.0 \\
 & 0.5 &  7 & \textbf{6.0} 				& 0.0&1.5 & \textbf{6} & 41.56 & \textbf{6} & {6} & 44.6 \\
 & 0.6 &  \textbf{4} & \textbf{4.0} 	& 0.0& 0.0 & \textbf{4} & 3.14& \textbf{4} & 4 & 9.8 \\ \hline

50 & 0.3 &  18 & \textbf{16} 			& 0.0& 226.4 & - & - & 0 & 15 & TL \\
 & 0.4 &  14 & \textbf{13.1} 			& 0.0& 117.4 & - & - & 0 & 11 & TL \\
 & 0.5 &  10 & \textbf{8.4} 			& 0.0& 128.7 & - & - & 0 & 7 & TL \\
 & 0.6 &  7 & \textbf{6} 				& 0.0& 87.7 & - & - & \textbf{6} & 6 & 1402.7 \\ \hline

100 & 0.3 &  {23} & \textbf{21.2} 		& 10.1& 461.6 & - & - & 0 & 13 & TL \\
 & 0.4 &  17 & \textbf{14.4} 			& 8.3& 393.5 & - & - & 0 & 10 & TL \\
 & 0.5 &  \textbf{12} & \textbf{12} 	& 0.0& 0.0 & - & - & 0 & 7 & TL \\
 & 0.6 &  {9} & \textbf{8.2} 			& 5.1&  222.7 & - & - & 0 & 6 & TL \\ \hline \hline

\end{tabular}}
\caption{Comparison of the approaches on benchmark set \textsc{Wireless} for $k=3$. } \label{tab:results-wireless-k-3}
\end{table}
	
	\begin{table}
	\centering
	\footnotesize{
		\begin{tabular}{rrrrrrlllll}
			\hline \hline
			\multicolumn{2}{c}{Instance} &
			\multicolumn{1}{c}{\textsc{Greedy}}   &
			\multicolumn{3}{c}{\textsc{VNS}}      &
			\multicolumn{2}{c}{\textsc{Ilp}}      &
			\multicolumn{3}{c}{\textsc{Benders}} \\
			
		     \cmidrule(rr){1-2}   \cmidrule(r){3-3} \cmidrule(rrr){4-6} \cmidrule(rr){7-8} \cmidrule(rrr){9-11}	   
			
			$n$ & $R$ &  obj & $\overline{obj}$ &  $\sigma[\%]$ & $\overline{t}_{best}[s]$ & obj & $t[s]$  & obj & dual & $t[s]$ \\ \hline
			
20 & 0.3 &  17 & \textbf{16}		&0.0 & 1.4 & \textbf{16} & 135.2 & \textbf{16} & 16 & 144.3 \\
 & 0.4 &  14 & \textbf{13} 			&0.0 & 49.1 & 19 & TL & 0 & 13 & TL \\
 & 0.5 &  8 & \textbf{7} 			&0.0 & 11.6 & 43 & TL & \textbf{7} & 7 & 267.9 \\
 & 0.6 &  \textbf{5} & \textbf{5} 	&0.0 & 0.0 & 46 & TL & \textbf{5} & 5 & 33.1 \\ \hline
50 & 0.3 &  \textbf{21} & \textbf{21}&2.8 & 0.0 & - & - & 0 & 15 & TL \\
 & 0.4 &  {16} & \textbf{15.6}		 & 0.0 & 18.7 & - & - & 0 & 12 & TL \\
 & 0.5 &  {12} & \textbf{11.1} 		 & 0.0 &  157.1 & - & - & 0 & 8 & TL \\
 & 0.6 &  8 & \textbf{7} 			& 0.0 & 16.9 & - & - & 0 & 7 & TL \\ \hline
 
100 & 0.3 & {28} & \textbf{27.4} 	& 0.0 & 261.5 & - & - & 0 & 14 & TL \\
 & 0.4 &  21 & \textbf{20} 			& 3.3 & 415.6 & - & - & 0 & 10 & TL \\
 & 0.5 &  \textbf{15} & \textbf{15} & 5.1 & 0.0  & - & - & 0 & 8 & TL \\
 & 0.6 &  {11} & \textbf{10.6} 		& 0.0 & 225.0 & - & - & 0 & 7 & TL \\
 \hline
\bottomrule
\end{tabular}}
\caption{Comparison of the approaches on benchmark set \textsc{Wireless} for $k=4$. } \label{tab:results-wireless-k-4}
\end{table}
	
	\subsubsection{Summary results on benchmark set \textsc{Real}}
	
	In order to show the efficiency of our \textsc{Vns} approach on benchmark set \textsc{Real},  the following methodology is applied. We take into a consideration those instances for which the both exact approaches were successful in solving them. Then,  the obtained solutions of \textsc{Greedy} and \textsc{Vns} approaches are compared with the exact solutions in terms of average solution qualities grouped by  $k \in \{2,   3,   4,   5\}$ (one row per each value $k$).  This is reported results  in Table~\ref{fig:real-comparisons-all}. Note that due to a high number of instances, \textsc{Vns} is executed only once per each instance. However,  due to a relatively small (relative) standard deviation obtained for the solutions of \textsc{Vns} on both benchmark sets,  \textsc{Random} and \textsc{Wireless},  it is not expected that overall conclusions   obtained for one-run \textsc{Vns} execution scenario will deviate significantly from the ten-runs \text{Vns} experimental scenario. \\

	\begin{table}
		\centering
		\begin{tabular}{ccccc} \hline
			$k$ & \textsc{Greedy} & \textsc{Vns}   & \textsc{Ilp}/\textsc{Benders} & \#instances \\ \hline
			2   & 20.98           & 20.15          & \textbf{20.11}                & 66         \\
			3   & 15.58           & \textbf{14.74} & \textbf{14.74}                & 43          \\
			4   & 13.96           & \textbf{13.00} & \textbf{13.00}                & 25          \\
			5   & 12.47           & \textbf{11.58} & \textbf{11.58}                & 17          \\ \hline \hline
		\end{tabular}
		\caption{Comparison between the four approaches on benchmark set \textsc{Real}: the case of the instances solved by \textsc{Ilp} and \textsc{Benders}.}
		\label{fig:real-comparisons-all}
	\end{table}
	The following conclusions may be drawn from there.
	
	\begin{itemize}
		\item The number of instances optimally solved is quickly deteriorating as $k$ increases.
		\item  \textsc{Vns} is able to find optimal solutions for all cases where \textsc{Ilp}/\textsc{Benders} were able to
		solve an instance (with an exception of a few instances for the case $k=2$).
		\item   \textsc{Greedy} approach is significantly  outperformed by the \textsc{Vns} approach.
		\item  We stress out that our two heuristic approaches are able to produce approximate solutions of reasonable quality (both are significantly better from the trivial solution) on the rest of the instances where the exact approaches could deliver none or only solutions of poor quality.
	\end{itemize}
	
	\subsection{Comparison between the two heuristic approaches}
\begin{figure}[H]
		\centering
		\begin{subfigure}{0.47\textwidth}
			\includegraphics[width=\linewidth,height=140pt]{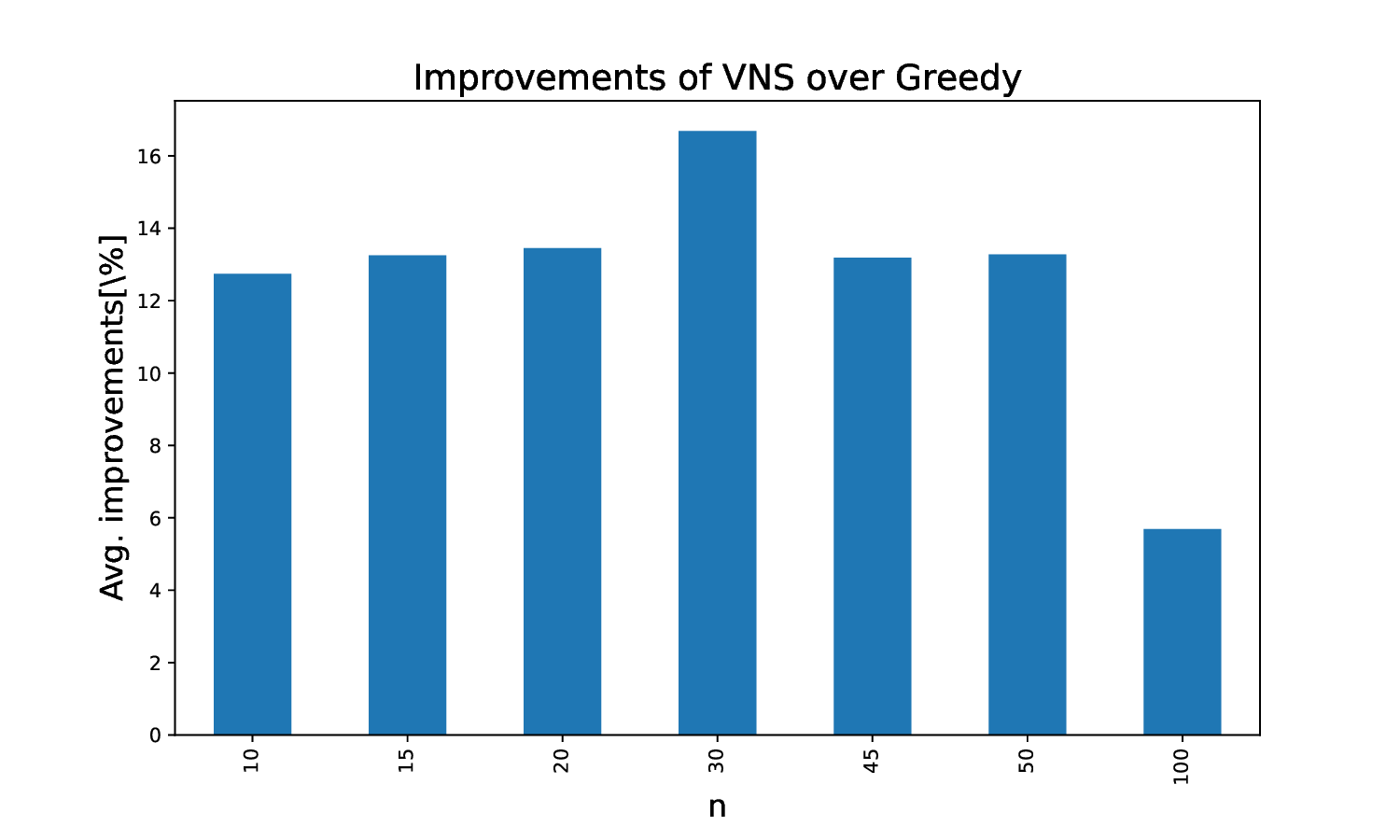}
			\subcaption{Instances grouped by $n$}
			\label{fig:improvements-plots-random-n}
		\end{subfigure}
		\hspace{0.3cm}
		\begin{subfigure}{0.47\textwidth}
			\includegraphics[width=\linewidth,height=140pt]{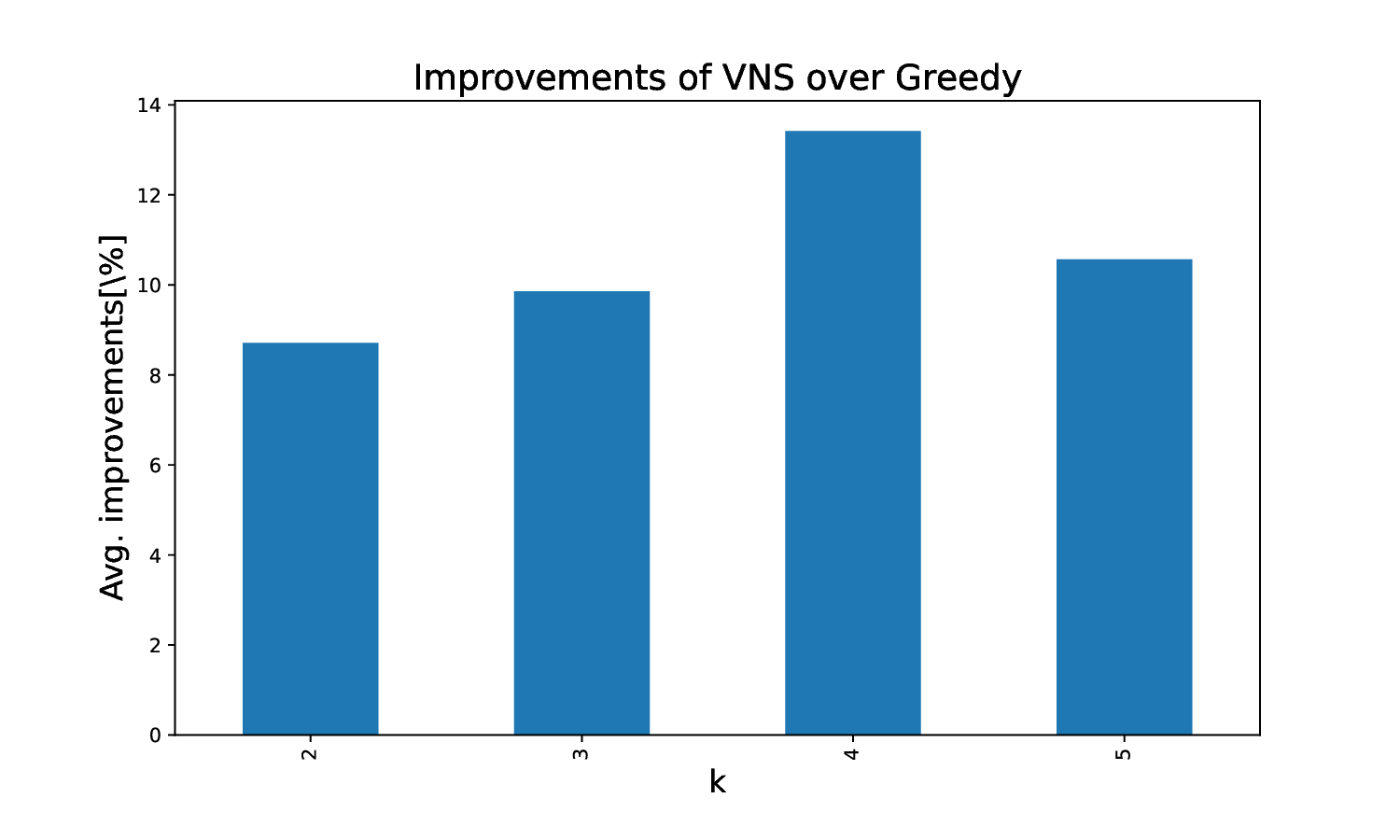}
			\subcaption{Instances grouped by $k$.}
			\label{fig:improvements-plots-random-k}
		\end{subfigure}
		\caption{Relative average improvements of \textsc{Vns} over \textsc{Greedy} approach on benchmark set \textsc{Random}.}
	\end{figure}

\begin{figure}[H]
		\centering
		\begin{subfigure}{0.47\textwidth}
			\includegraphics[width=\linewidth,height=140pt]{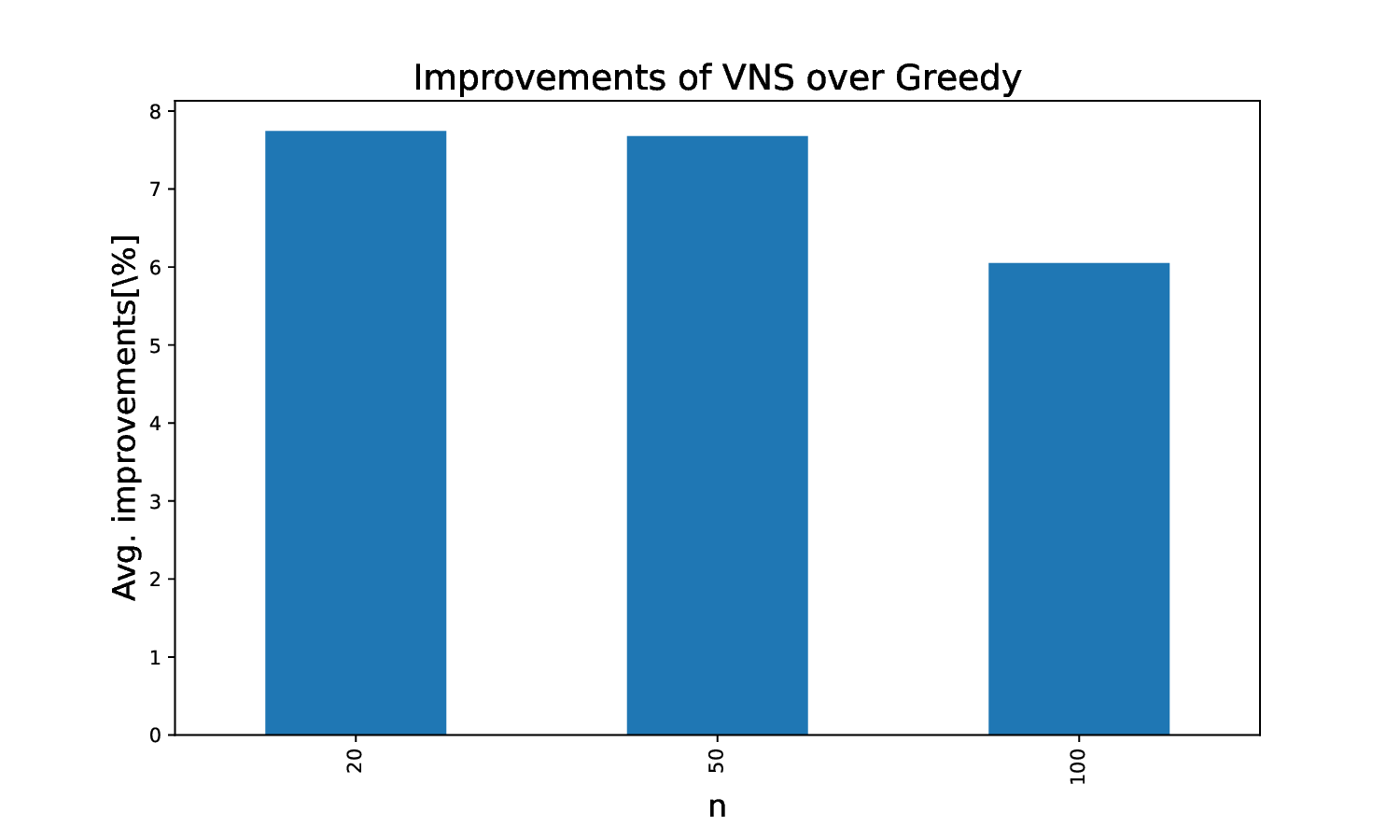}
			\subcaption{Instances grouped by $n$}
			\label{fig:improvements-plots-wireless-n}
		\end{subfigure}
		\hspace{0.3cm}
		\begin{subfigure}{0.47\textwidth}
			\includegraphics[width=\linewidth,height=140pt]{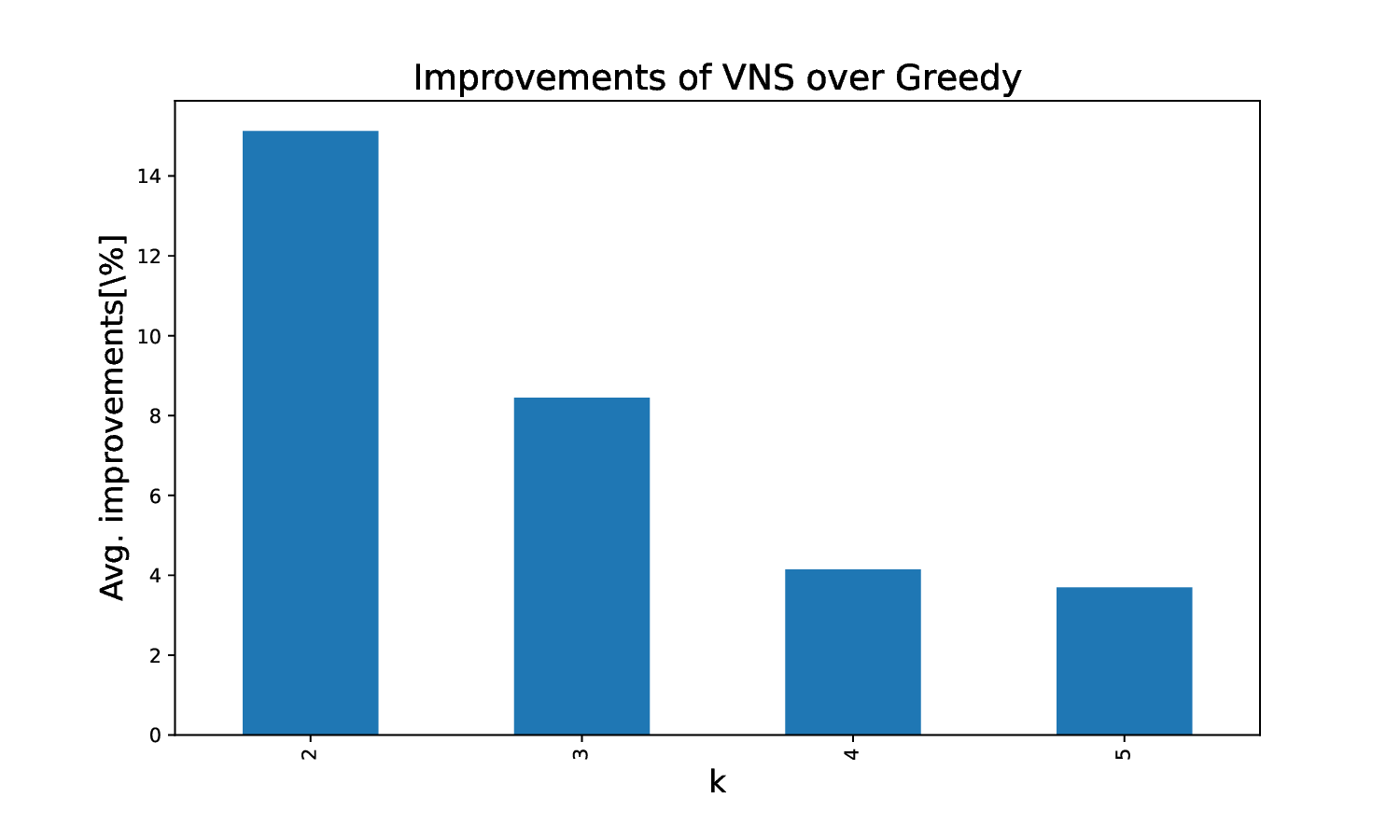}
			\subcaption{Instances grouped by $k$.}
			\label{fig:improvements-plots-wireless-k}
		\end{subfigure}
		\caption{Relative average improvements of \textsc{Vns} over \textsc{Greedy} approach on benchmark set \textsc{Wireless}.}
	\end{figure}

\begin{figure}[H]
		\centering
		\includegraphics[width=230pt,height=140pt]{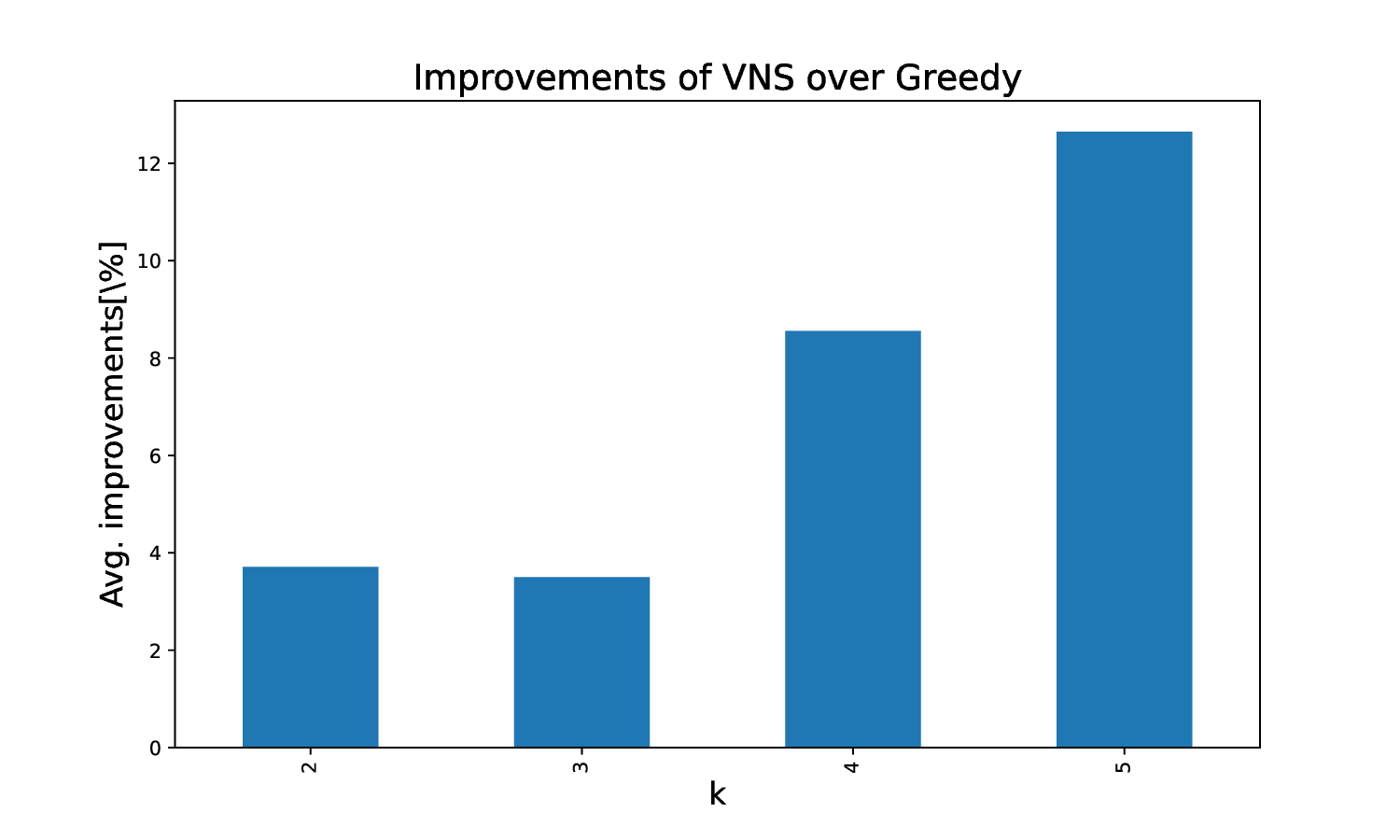}
		\caption{Relative average improvements of \textsc{Vns} over \textsc{Greedy} approach on benchmark set \textsc{Real}.}
		\label{fig:improvements-plots-real-k}
	\end{figure}

	To compare the two of our heuristic approaches,   the \textsc{Greedy} and \textsc{Vns},  we have additionally provided Figure~\ref{fig:improvements-plots-random-n}--\ref{fig:improvements-plots-random-k} for benchmark set \textsc{Random} and Figure~\ref{fig:improvements-plots-wireless-n}--\ref{fig:improvements-plots-wireless-k} for benchmark set \textsc{Wireless}. These plots display the average relative improvements of \textsc{Vns} over \textsc{Greedy} in terms of solution quality. The instances are grouped by $n$ and by $k$ (the values shown on $x$-axis) separately; thus, two plots are generated per each benchmark set. Additionally Figure~\ref{fig:improvements-plots-real-k} displays the average relative improvements of \textsc{Vns} over \textsc{Greedy} in terms of solution quality for all instances from benchmark set \textsc{Real} where the instances are grouped by $k$.

	
	The following conclusions may be drawn from generated plots.
	\begin{itemize}
		\item For the instances from the benchmark set \textsc{Random} grouped by graph dimension ($n$),  average improvements of the quality of initial solutions of \textsc{Vns} range from respectable 12-16\% for $10 \leq n \leq 50$ whereas for the largest group of instances with $n=100$ this percentage still remains significant, and is at about 5\%.
		\item  For the instances from benchmark set \textsc{Random} grouped by  $k$ values, average relative improvements of \textsc{Vns} range from 8.5\% for $k=2$ to 13\%  for the group with $k=4$.
		\item  For the instances from the benchmark set \textsc{Wireless} grouped by $n$, the average improvements of \textsc{Vns} range from about 6\% for the largest $n=100$ and $\approx$7.7\% for smaller $n \in \{20,  50\}$.
		\item  For the instances from the benchmark set \textsc{Wireless} grouped by $k$, the average improvements expectedly deteriorate with the increase of $k$ value. For the smallest $k=2$ the average improvements are about an excellent 15\% and decrease for the largest  case $k=5$  to 3\%. This is mainly due to a high number of generated attacks that have to be verified for the existence of a defense strategy which significantly reduces the overall number of performed iterations of the \textsc{Vns} approach.
		\item  For the instances from the benchmark set \textsc{Real}, average relative improvements of \textsc{Vns} range from 3.5\%--3.8\% for $k\in\{2,3\}$ to more than  12\% for the largest case  $k=5$.
	\end{itemize}
	
	\subsection{Statistical analysis}
	
	To verify the statistical significance of the observed differences between the four approaches,  we performed the following statistical methodology.
	First, Friedman’s test was executed for all four competitor approaches on all problem instances of the \textsc{Random} and \textsc{Wireless} benchmark sets separately.
	After that, in cases when the null hypothesis was rejected ($H_0$ says that competitor approaches are statistically equally good) pairwise comparisons were further performed by using the Nemenyi post-hoc test~\cite{pohlert2014pairwise}.

	The outcome is represented within critical difference (CD) plots. A CD plot positions each approach on the horizontal axis according to its average ranking. Then, the CD is computed for a significance level of 0.05. If the difference is small enough, that is, no statistical difference is detected, a horizontal bar connecting statistically equal approaches is drawn. In case no result has been returned by an algorithm, we simply assign a large value (e.g. 1000) providing with the largest possible ranking.

	\begin{figure}[H]
		\centering
		\begin{subfigure}{0.45\textwidth}
			\includegraphics[width=\linewidth]{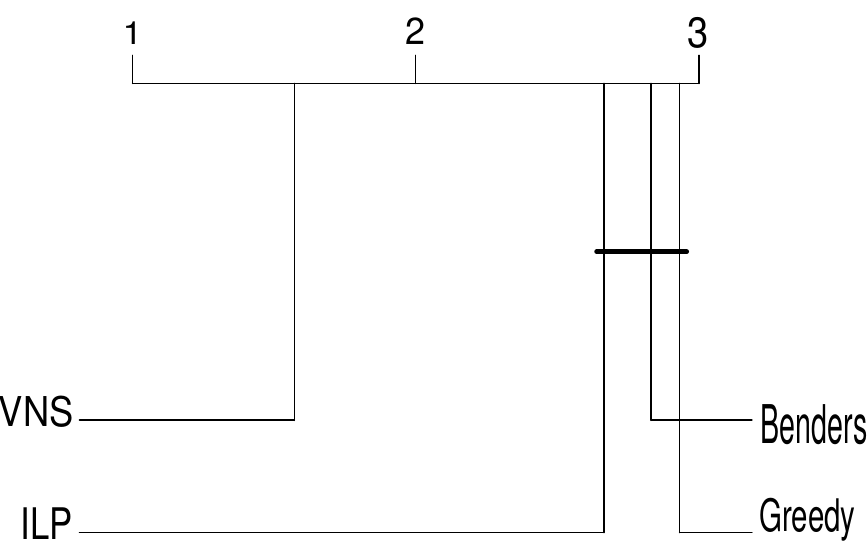}
			\subcaption{Benchmark set \textsc{Random}}
			\label{fig:cd-plots-random}
		\end{subfigure}
		\hspace{0.5cm}
		\begin{subfigure}{0.45\textwidth}
			\includegraphics[width=\linewidth]{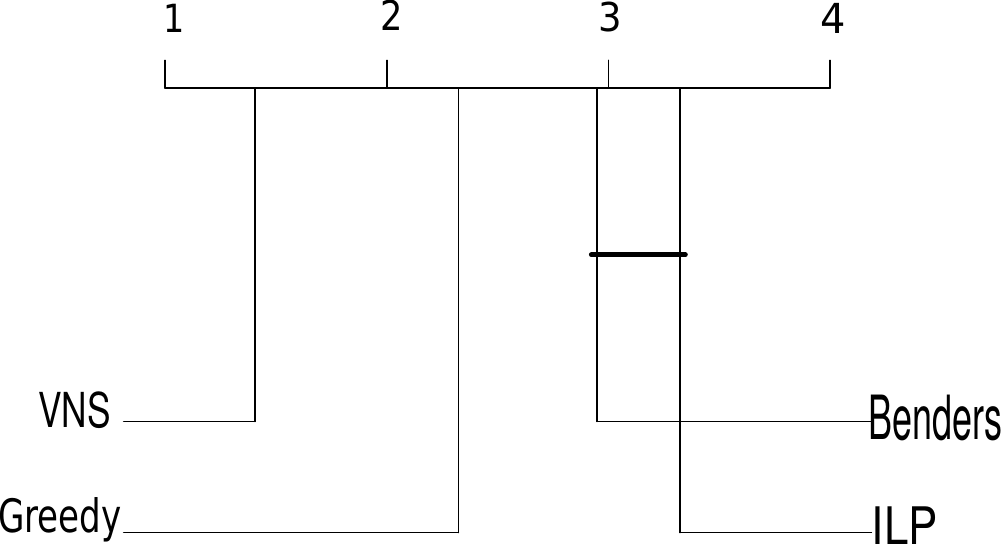}
			\subcaption{Benchmark set \textsc{Wireless}}
			\label{fig:cd-plots-wireless}
		\end{subfigure}
		
		\begin{subfigure}{0.45\textwidth}
			\includegraphics[width=\linewidth]{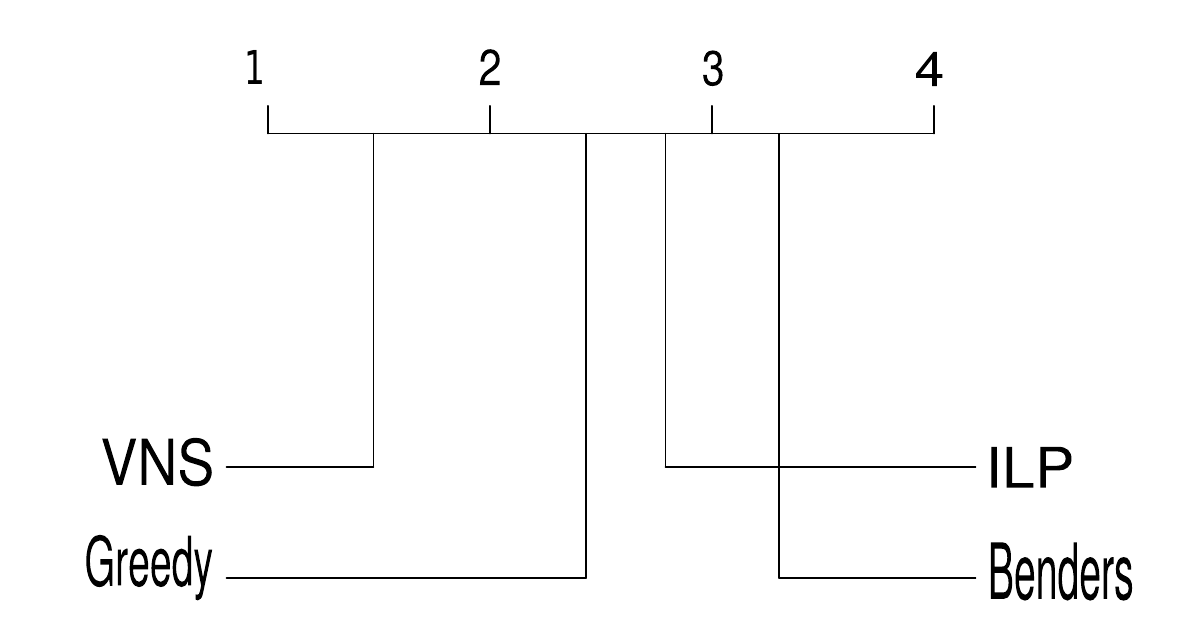}
			\subcaption{Benchmark set \textsc{Real}}
			\label{fig:cd-plots-real}
		\end{subfigure}
		\caption{CD plots showing statistical differences between the four approaches on the results summarized over all $k\in\{2,3,4,5\}$.}
	\end{figure}

	The following conclusions may be drawn from the statistical analysis conducted, as depicted in Figures~\ref{fig:cd-plots-random}--		\ref{fig:cd-plots-real}.
	\begin{itemize}
		\item For the benchmark set \textsc{Random}, \textsc{Vns} approach achieves the best average ranking while the other three approaches are far behind, obtaining the average ranking that is close to the value of three.  Additionally, \textsc{Vns} is performing statistically significantly better from the remainder of the competitors and between three of them no statistically significant difference occurred.
		\item For the benchmark set \textsc{Wireless}, \textsc{Vns} again gains the best average ranking, which is for the whole ranking of one better than the second best \textsc{Greedy} approach. The worst average ranking is delivered by \textsc{Benders} and \textsc{Ilp}, in this order. Again, the best performing approach is \textsc{Vns}, which significantly outperforms the \textsc{Greedy} approach in terms of obtained solution quality. Statistically the worst performing approaches are \textsc{Benders} and \textsc{ILP} as they could not usually deliver any solution for larger $k$ values.
		\item For the benchmark set \textsc{Real}, \textsc{Vns} again obtains the best average ranking. The second best average ranking is achieved by \textsc{Greedy} approach, primarily due to its consistent delivery of feasible solutions, unlike the two exact competitor algorithms. Note that again the solutions of \textsc{Vns} are statistically significantly better than that of the \textsc{Greedy} approach.
	\end{itemize}

	\section{Application of the $k$-SRD problem to a real-world scenario} \label{sec:case-study}
	
	In the pursuit for applicability, we decided to visualize and interpret the results of \textsc{Vns} approach on a few real-world instances generated from GeoJSON files, see Section~\ref{sec:experimental-evaluation}. 
	For a real-world case scenario we opt for a location problem which concern placing fire stations with a number of fire trucks in a city at appropriate locations under specific constraints.  As follows we list some numbers, mainly taken from Wikipedia, as the reference point for this real-world problem; e.g.  the number of fire stations in the city of Dublin is 14, in the city of Berlin is 35, in Chicago there are 98 fire stations, while in the city of New York, 254 stations are installed.  
	The number of available fire trucks  in Chicago is 61, and in the city of New York is 143. In the city of Berlin,  in 2019, its Fire Defense received 478,281 emergency calls, known as the busiest of all fire services in Germany. Approximately 83\% of the alarms per year are for the emergency services, 5\% for technical assistance and only about 2\% for firefighting\footnote{This information is phrased at the link \url{https://en.wikipedia.org/wiki/Berlin\_Fire\_Brigade}}.
	If we assume that 4 hrs are needed on average for a complete (successful) intervention (from exit of the fire engine from the station to its re-entry after the intervention and its re-use), the number of  real fire emergencies that may simultaneously occur in this city is expected to be $\approx k=5$.
	
	Let us find the best decisions of installing fire stations at certain districts under the following (simplified) constraints.
	\begin{itemize}
		\item At most one fire station might be  built per each district.
		\item At any chosen location for building a fire stations, one or more fire trucks can be placed.
		\item One fire truck placed at a fire station of a district can serve to extinguish fire occuring in that district or the neighboring districts. This reflects on realistic scenario of demanding  quick  speed of reaction in this kind of emergency. For example, sending a fire truck from the westernmost part of the city to serve in a district diametrically opposite will consume valuable time, particularly in big cities with large diameters and expected heavy traffic congestion.
		\item At least one truck is always kept to secure that district, i.e. all but one fire truck at any station are available to be sent to neighboring districts for interventions.
	\end{itemize}
	The given value of $k$ represents an estimator for the number of districts that can be simultaneously under fire. We additionally assume a simplified scenario that one fire truck is sufficient for extinguishing any fire (may not be that realistic in a huge threat).  The aim is to find locations to build fire stations along with the number of fire trucks in each of them so that any fire scenario that occurs in $k$ different districts simultaneously, can be treated safely.
	
	We demonstrate visually applications of the above problem on a few world cities whose graph representations are extracted from corresponding GeoJSON files, see Section~\ref{sec:experimental-evaluation}. Three cities are chosen: Szczecin and Cincinnati as the representatives for medium-sized instances and Pittsburgh as a representative of large-sized instance. The value of $k $ is chosen from the set $\{5,6,8\}$. The city of Szczecin comprises 36 districts into one connected graph. The city of Cincinnaty consists of 45 districts, also forming a connected graph. On the other hand, the city of Pittsburgh consists of  90 districts subdivided into 4 disjoint regions.  Note that none of the exact approaches could successfully solve any of these instances, and the solutions provided by VNS consistently exhibit higher quality than those of the \textsc{Greedy} approach. 
	
\begin{figure}[H]
		\centering
		\begin{subfigure}{0.47\textwidth}
			\includegraphics[width=\linewidth]{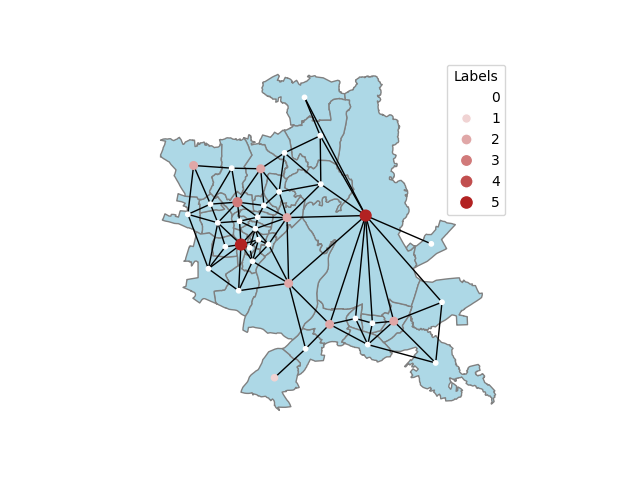}
			\subcaption{$k=5$}
			\label{fig:szcz-5}
		\end{subfigure}
		\hspace{0.5cm}
		\begin{subfigure}{0.47\textwidth}
			\includegraphics[width=\linewidth]{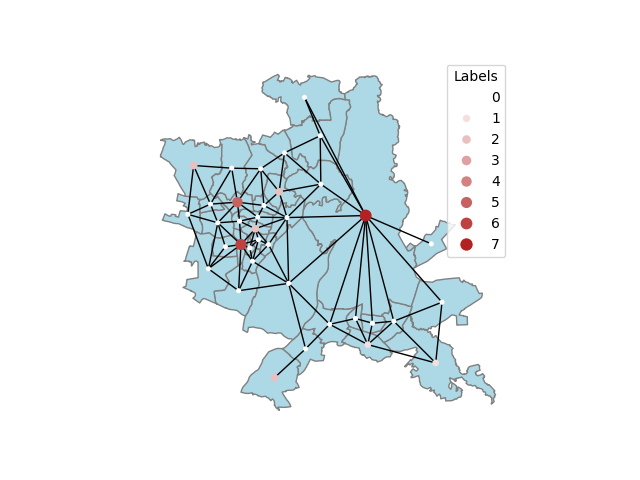}
			\subcaption{$k=6$}
			\label{fig:szcz-6}
		\end{subfigure}
		
		\begin{subfigure}{0.50\textwidth}
			\includegraphics[width=\linewidth]{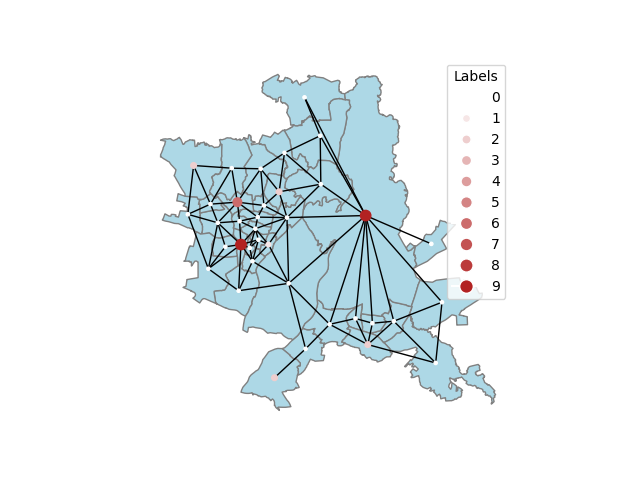}
			\subcaption{$k=8$}
			\label{fig:szcz-8}
		\end{subfigure}
		\caption{Graph of the city of Szczecin.}
	\end{figure}

\begin{figure}[H]
		\centering
			\begin{subfigure}{0.47\textwidth}
				\includegraphics[width=\linewidth,height=120pt]{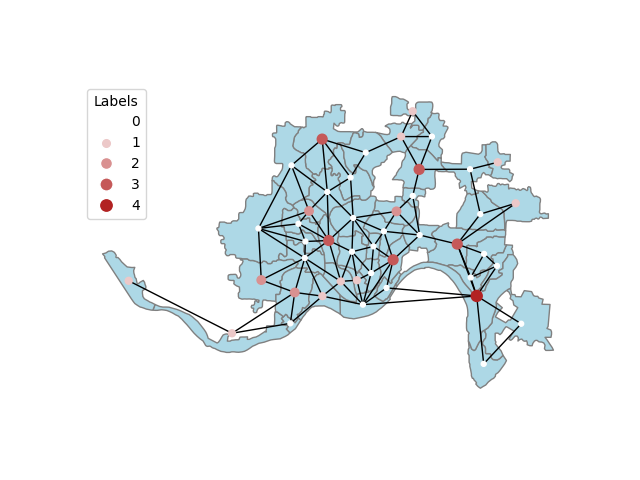}
				\subcaption{$k=5$}
				\label{fig:cincinnaty-5}
			\end{subfigure}
			\hspace{0.2cm}
			\begin{subfigure}{0.47\textwidth}
				\includegraphics[width=\linewidth,height=120pt]{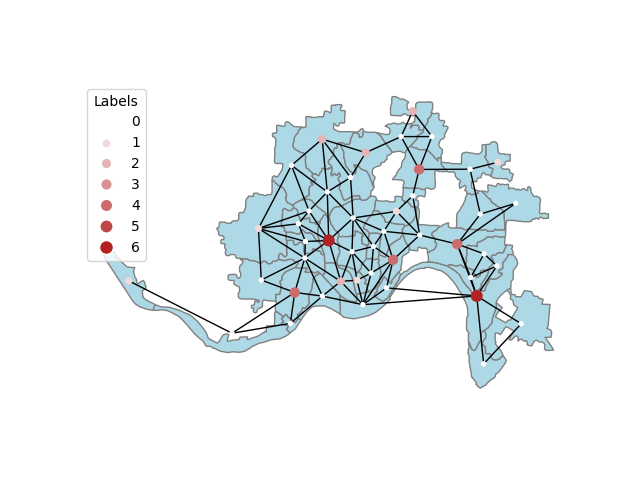}
				\subcaption{$k=6$}
				\label{fig:cincinnaty-6}
			\end{subfigure}
			
			\begin{subfigure}{0.47\textwidth}
				\includegraphics[width=\linewidth,height=120pt]{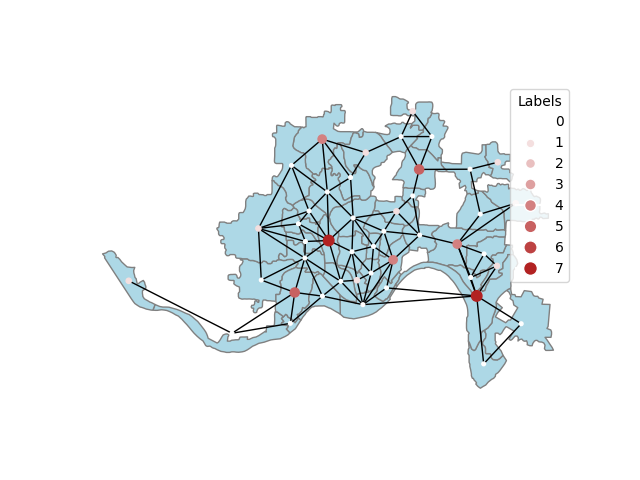}
				\subcaption{$k=8$}
				\label{fig:cincinnaty-8}
			\end{subfigure}
		\caption{Graph of the city of Cincinnati.}
	\end{figure}

\begin{figure}[H]
		\centering
			\begin{subfigure}{0.47\textwidth}
				\includegraphics[width=\linewidth,height=130pt]{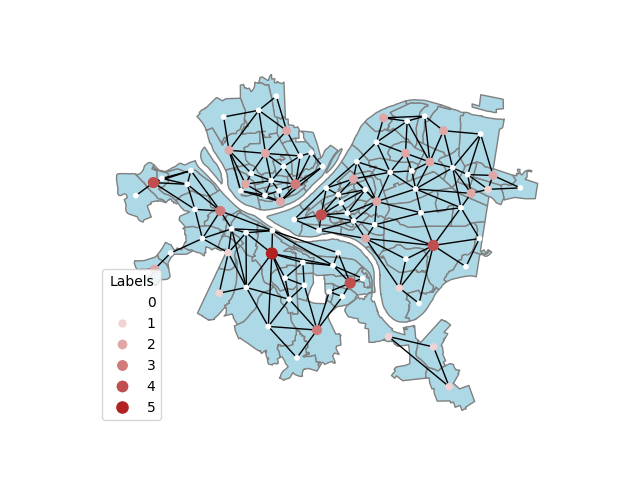}
				\subcaption{$k=5$}
				\label{fig:pittsburgh-5}
			\end{subfigure}
			\hspace{0.2cm}
			\begin{subfigure}{0.47\textwidth}
				\includegraphics[width=\linewidth,height=130pt]{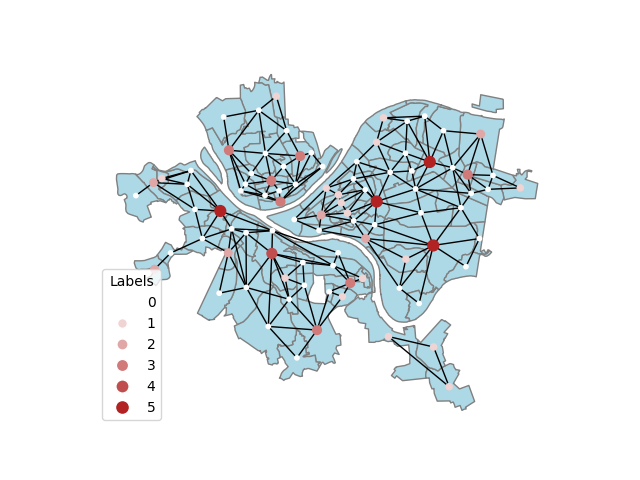}
				\subcaption{$k=6$}
				\label{fig:pittsburgh-6}
			\end{subfigure}
			
			\begin{subfigure}{0.47\textwidth}
				\includegraphics[width=\linewidth,height=130pt]{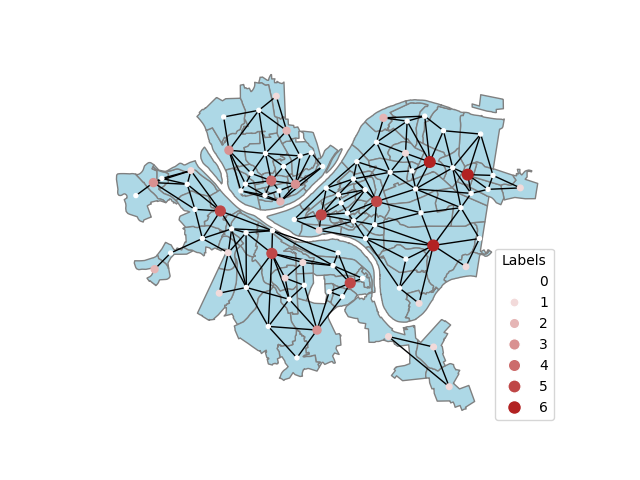}
				\subcaption{$k=8$}
				\label{fig:pittsburgh-8}
			\end{subfigure}
		\caption{Graph of the city of Pittsburgh.}
	\end{figure}

	The solutions produced by \textsc{Vns} for all three cities (and all three values of $k$) are displayed in Figures~\ref{fig:szcz-5}--\ref{fig:pittsburgh-8}.   The following conclusions are made for the city of Szczecin.
	\begin{itemize}
		
		\item For $k=5$ the solution is equal to 26.  10 districts are chosen as hubs for building fire stations where at least one fire truck is placed.  There are two centralized hubs that each keep five fire trucks and one that keeps three trucks.  There are six fire stations with two trucks and one with one truck. For the two hubs with dominant number of trucks, one is located in the geographically largest district, and the second is among the smallest districts, close to the city center whose reachability is high.
		\item For $k=6$, the solution is equal to 28. Nine fire stations are supposed to be built. One station keeps seven trucks, one with six trucks and one with five trucks.  The others keep two or less than two trucks.
		\item For $k=8$, the obtained solution  is equal to 33. Eight fire stations are suggested to be built. Again, three major stations are distinguishable in terms of the number of trucks.  Two of them keep nine trucks, and one comes with six trucks. The others keep two or less than two trucks.
		
		\item Interestingly, two locations for fire stations with a large number of trucks  match for any  $k\in\{5, 6, 8\}$.  For $k\in\{6, 8\}$, three of the locations match.
		This can be argued by the fact that the corresponding nodes have the highest centrality degree among most other nodes, indicating their high immediate reachability. Thus, they can cover many surrounding districts if enough trucks are placed. The two common nodes for each value of $k$ have degree centrality score of  0.25 and 0.31, and are among the highest for all nodes.
		
	\end{itemize}

	The following conclusions are drawn for the graph of the city of Cincinnati.
	
	\begin{itemize}
		\item For $k=5$, the obtained solution is equal to 36. 19 fire stations are suggested to open. There is no station that keeps more than four fire trucks. There is just one station keeping 4 trucks and many (5) of them with three. The stations with more fire trucks seem to be geographically uniformly distributed.
		\item For $k=6$, the obtained solution is equal to 41. 15 fire stations are suggested to be built. There are two stations with six fire trucks and four stations with four fire trucks while the remaining stations comprise two or less fire trucks.
		\item For $k=8$, the obtained solution is equal to 45. 16 fire stations are suggested to be built. Two fire stations keep seven fire trucks, two stations with five, and three with four fire trucks, while for  nine stations just one truck is assigned (mainly located at the boundary of the whole city region).  The seven stations that dominate in term of number of kept trucks seem to be geographically equally distributed over the city region. 
		\item The two nodes assigned with the largest number of trucks  (7) have the associated highest centrality degree scores (0.18 and 0.16) in the graph.
	\end{itemize}

	Last but not least, we visualize and analyze the solution of \textsc{Vns} for the city of Pittsburgh. The following facts are observed.   
	
	\begin{itemize}
		\item For $k=5$, the obtained solution is equal to 67. There are 30 locations suggested for a hub, i.e. to build a fire station. There is one station keeping five fire trucks, and three with four fire trucks. Three stations also hold three fire trucks, and the rest of them store two or less trucks.

		\item For $k=6$, the obtained solution is equal to 73. There are 32 locations chosen for building fire stations. Four of them come with five fire trucks, one with four and seven of them with three  trucks.
		\item For $k=8$ the obtained solution is equal to 81.  There are 31 locations suggested for building a fire station. Three of them keep six fire trucks, five of them keep five trucks, and four of them three trucks.
		\item While for $k=6$ and $k=8$, the locations of stations with the highest number of trucks mostly match, for the case $k=5$ it is less clear; however, there exist two locations which correspond to the locations with a larger number of trucks in case of both $k=6$ and $k=8$.
		\item Concerning the mentioned two common dominant locations (nodes) in terms of number of trucks located for each $k \in \{5,6,8\}$,  their centrality degrees are among the highest in the graph and are equal to 0.09 and 0.06. 
	\end{itemize}


	\section{Conclusions and future work}\label{sec:conclusion}

	In this paper we have dealt with solving the $k$-strong Roman domination ($k$-SRD) problem with arbitrary $k\geq 2, k \in \mathbb{N}$. The $k$-SRD problem is extremely hard to solve, since even verifying the feasibility of a solution is exponential with the number of nodes in graph. We have proposed two heuristic approaches: ($i$) a greedy approach based on node coverage information, and ($ii$) an improved variable neighborhood search (VNS) algorithm that uses the greedy solution as the initial solution. Our VNS approach is based on three basic principles. First, a shaking procedure that shakes components of the best solution based on roulette selection by always generating a one better but possibly infeasible solution. Second, relaxed feasibility is checked during the search; this concept is called quasi-feasibility. Here, a  set of $k$-attacks in the vicinity of the   nodes of the considered solution is carefully generated. Then, for all these attacks, a strategy is probed for defending them. Two strategies are used for defense -- one deterministic and one probabilistic (roulette-based). The fitness score of the solution is made up of two components: the number of predefined attacks that cannot be defended and the sum of all labels of the given solution. The third core principle of VNS is an efficient local search procedure based on the 2-decomposition operator on a swap basis, which includes an initial improvement strategy and an internal caching mechanism to ensure efficient execution.
	
	Two of our heuristic approaches are compared with the two existing exact approaches from the literature, integer linear programming (ILP) and Benders-based approaches. This comparison is performed on randomly generated graphs from the literature as well as on two newly introduced benchmark sets: wireless ad hoc networks with different densities and real-word instances generated from corresponding GeoJSON files. These files represent spatial data of cities divided into districts, countries into regions, etc. While the two exact approaches could only handle small to medium sized instances and some slightly larger ones with $k=2$, they were not able to handle the largest instances as they consistently encountered  issues with memory (ILP) or time (the Benders-based approach).
	Our VNS approach was able to consistently and significantly improve the results of the greedy approach on instances of different sizes and distributions. Moreover, for most instances solved by some of the exact approaches, our VNS approach was able to achieve these solutions or produce near-optimal, statistically equally good solutions in reasonable time. For the real-world instances and their solutions obtained from solving the $k$-SRD problem by the VNS approach for larger, and never so-far tested  values of $k\in\{6,8\}$, a case study of deciding on opening locations for fire stations in the neighbourhoods and assigning a number of fire trucks to each of these locations to be able to handle at most $k$ simultaneous fires is discussed.
	
	For future work, one could design an approach for much larger problem instances with thousands of nodes. Note that our VNS approach might be limited to an application of graphs with up to a few hundred nodes; the number of predefined attacks grows rapidly with the size of the graph, making the approach computationally expensive per iteration. One idea is to transform the current VNS into a more scalable VNS by intelligently clustering a graph into some dependent groups of nodes, e.g. using community detection algorithms. Then, the standalone VNS would be applied to each of the obtained components (one could also develop a parallel implementation). The combination of the obtained solutions for each of the subgraphs induced by the clusters into a high quality solution of the initial graph could be done by a local search based procedure for detected critical nodes. Another aspect could be to design a different meta-heuristic approach that builds on our greedy approach by randomizing it and improving the generated solution e.g. by the established appropriate local search procedure. \\ \vspace{0.3cm}
	
	\textbf{Acknowledgments}.   Marko Djukanovic, Stefan Kapunac, and Dragan Matic acknowledge are partially funded by the project entitled ''Developing models and algorithms of Artificial intelligence for solving computationally hard combinatorial optimization problems'' under no. 1259086 supported by the Ministry of Scientific and Technological Development and Higher Education of Republic of Srpska.  Marko Djukanovic is additionally supported by the project grant ''Theoretical and computational aspects of solving Roman domination problems'' funded by the Ministry of Civil Affairs of Bosnia and Herzegovina under no. 1259074.

	\newpage
	\appendix
	
	\section{Pseudocodes supporting is\_attack\_defended() function} \label{app: pseudocodes}
	
	\begin{algorithm}[H]
		\footnotesize{
			\begin{algorithmic}[1]
				\Statex \textbf{Input}: solution $s$, graph $G=(V,E), k, all\_alternatives, reduced\_attack, defending\_nodes$
				\Statex \textbf{Output}: a pair \emph{attack\_defended}, \emph{defending\_nodes}
				\For{$alternative \in all\_alternatives$}
				\State $attack\_defended \gets True$
				\For{$v \in reduced\_attack$}
				\State $u \gets alternative[v]$
				\If{$u$ can help $v$}
				\State add $u$ to $defending\_nodes$
				\Else
				\State $attack\_defended \gets False$
				\State \textbf{break}
				\EndIf
				\EndFor
				\If{$attack\_defended$}
				\State \textbf{return} $attack\_defended, defending\_nodes$
				\EndIf
				\EndFor
				\State \textbf{return} $False, \emptyset$
				
		\end{algorithmic}}
		\caption{Function \texttt{deterministic\_defense()}}
		\label{alg:determinitic-defense}
	\end{algorithm}
	\begin{algorithm}[H]
		\footnotesize{
			\begin{algorithmic}[1]
				\Statex \textbf{Input}: solution $s$, graph $ G=(V, E), k, reduced\_attack, defending\_nodes, tries>0$
				\Statex \textbf{Output}: a pair \emph{attack\_defended}, \emph{defending\_nodes}
				
				\While{$tries > 0$}
				\State $attack\_defended \gets True$
				\State $ tries \gets tries - 1$
				\For{$v \in reduced\_attack$}
				\State $candidates \gets []$
				\For{$u \in N(v)$}
				\State $diff \gets$ number of available FAs in node $u$
				\State $candidates.add([\underbrace{u, \ldots, u}_{\emph{diff} \textrm{ times}}])$
				\EndFor
				\If{$candidates.size() = 0$}
				\State $attack\_defended \gets False$
				\State \textbf{break}
				\EndIf
				\State $selected\_u \gets$ random element from $candidates$
				\State add $selected\_u$ to $defending\_nodes$
				\EndFor
				\If{$attack\_defended$}
				\State \textbf{return} $attack\_defended, defending\_nodes$
				\EndIf
				\EndWhile
				\State \textbf{return} $False, \emptyset$
				
		\end{algorithmic}}
		\caption{Function \texttt{roulette\_defense()}}
		\label{alg:probabilistic-defense}
	\end{algorithm}

	\section{Complete  results for benchmark set \textsc{Random}}\label{app: random-remaining-results}
	
	\begin{table}[H]
	\caption{Comparison of the approaches on benchmark set \textsc{Random} for $n=10$. } \label{tab:results-random-n-10}
	 \centering
	\footnotesize{
		\begin{tabular}{rrrrrrrlllll}
			\hline \hline
			\multicolumn{2}{c}{Instance} &
			\multicolumn{1}{c}{$k$} &
			\multicolumn{1}{c}{\textsc{Greedy}}   &
			\multicolumn{3}{c}{\textsc{VNS}}      &
			\multicolumn{2}{c}{\textsc{Ilp}}      &
			\multicolumn{3}{c}{\textsc{Benders}} \\ 
		\cmidrule(rr){1-2}   \cmidrule(r){4-4} \cmidrule(rrr){5-7} \cmidrule(rr){8-9} \cmidrule(rrr){10-12}	 
		$n$ & \textit{index} & & obj & $\overline{obj}$ &  $\sigma[\%]$ & $\overline{t}_{best}[s]$ & obj & $t[s]$ & obj & dual &  $t[s]$ \\ \midrule
		
10 & 1 & 2 & 7 & \textbf{6}		&0.0 & 0.0 & \textbf{6} & 0.1 & \textbf{6} & 6 & 0.7 \\
 & 2 &  & \textbf{6} & \textbf{6}			&0.0 & 0.0 & \textbf{6} & 0.1 & \textbf{6} & 6 & 0.7 \\
 & 3 &  & \textbf{6} & \textbf{6}			&0.0 & 0.0 & \textbf{6} & 0.1  & \textbf{6} & 6 & 2.5 \\
 & 4 &  & 6 & \textbf{5}			&0.0 & 0.0 & \textbf{5} & 0.0  & \textbf{5} & 5 & 1.2 \\
 & 5 &  & \textbf{6} & \textbf{6}			&0.0 & 0.0 & \textbf{6} & 0.1 & \textbf{6} & 6 & 1.8 \\ \hline
10 & 1 & 3 & 8	&\textbf{7}      & 0.0 & 0.0 & \textbf{7} & 0.5  & \textbf{7} & 7 & 2.4 \\
 & 2 &  & 7 & \textbf{6}			&0.0 & 0.0 & \textbf{6} & 0.4  & \textbf{6} & 6 & 3.8 \\
 & 3 &  & \textbf{7} & \textbf{7}			& 0.0& 0.0 & \textbf{7} & 0.4  & \textbf{7} & 7 & 3.0 \\
 & 4 &  & 8 & \textbf{6}			& 0.0& 0.1 & \textbf{6} & 0.3 & \textbf{6} & 6 & 2.9 \\
 & 5 &  & \textbf{7} & \textbf{7}			&0.0 & 0.0 & \textbf{7} & 0.4  & \textbf{7} & 7 & 2.9\\ \hline
10 & 1 & 4 & 9 & \textbf{7}		&0.0 & 0.4 & \textbf{7} & 1.6 & \textbf{7} & 7 & 3.4 \\
 & 2 &  & 8 & \textbf{7} 		&0.0 & 0.0 & \textbf{7} & 1.6 & \textbf{7} & 7 & 3.4 \\
 & 3 &  & 8 & \textbf{7}			&0.0 & 0.5 & \textbf{7} & 1.4 & \textbf{7} & \textbf{7} & 4.1 \\
 & 4 &  & 9 & \textbf{7}			&0.0 & 0.4 & \textbf{7} & 1.3  & \textbf{7} & \textbf{7} & 4.2 \\
 & 5 &  & 8 & \textbf{7}			&0.0 & 0.1 & \textbf{7} & 1.4 & \textbf{7} & 7 & 4.4 \\ \hline
10 & 1 & 5 & 10 & \textbf{8}		&0.0 & 0.1 & \textbf{8} & 4.1 & \textbf{8} & 8 & 6.3 \\
 & 2 &  & 9 & \textbf{8}         & 0.0& 0.0 & \textbf{8} & 2.8 & \textbf{8} & 8 & 4.4\\
 & 3 &  & 9 & \textbf{8}			& 0.0& 0.0 & \textbf{8} & 3.2 & \textbf{8} & 8 & 11.5 \\
 & 4 &  & 10 & \textbf{8}		&0.0 & 0.0 & \textbf{8} & 2.7  & \textbf{8} & 8 & 7.1 \\
 & 5 &  & 9 & \textbf{8}			&0.0 & 0.2 & \textbf{8} & 2.0 & \textbf{8} & 8 & 9.7 \\ \hline \hline
\end{tabular}}

\end{table}

	\begin{table}[H]
	\caption{Comparison of the approaches on benchmark set \textsc{Random} for $n=15$. } \label{tab:results-random-n-15}
	\centering
	\footnotesize{
		\begin{tabular}{rrrrrrrlllll}
			\hline \hline
			\multicolumn{2}{c}{Instance} &
			\multicolumn{1}{c}{$k$} &
			\multicolumn{1}{c}{\textsc{Greedy}}   &
			\multicolumn{3}{c}{\textsc{VNS}}      &
			\multicolumn{2}{c}{\textsc{Ilp}}      &
			\multicolumn{3}{c}{\textsc{Benders}} \\
		    
		    \cmidrule(rr){1-2}   \cmidrule(r){4-4} \cmidrule(rrr){5-7} \cmidrule(rr){8-9} \cmidrule(rrr){10-12}	 
			
			$n$ & \textit{index} & & obj & $\overline{obj}$ &  $\sigma[\%]$ & $\overline{t}_{best}[s]$ & obj & $t[s]$ & obj & dual &  $t[s]$ \\ \hline
15 & 1 & 2 & 10 & \textbf{9} & 	0.0	& 0.0 & \textbf{9} & 0.9 & \textbf{9} & 9 & 5.3 \\
 & 2 &  & \textbf{8} & \textbf{8} &		0.0	& 0.0 & \textbf{8} & 0.2 & \textbf{8} & 8 & 2.2 \\
 & 3 &  & 9 & \textbf{8} &		0.0	& 0.0 & \textbf{8} & 0.2 & \textbf{8} & 8 & 2.8  \\
 & 4 &  & 10 & \textbf{9}&		0.0	& 0.0 & \textbf{9} & 1.2  & \textbf{9} & 9 & 5.0 \\
 & 5 &  & 10 & \textbf{9}& 		0.0	& 0.2 & \textbf{9} & 0.3 & \textbf{9} & 9 & 3.8  \\ \hline
15 & 1 & 3 & 12 & \textbf{10} &	0.0	& 0.3 & \textbf{10} & 12.2  & \textbf{10} & 10 & 20.2  \\
 & 2 &  & \textbf{10} & \textbf{10}& 		0.0	& 0.0 & \textbf{10} & 3.6 & \textbf{10} & 10 & 12.9  \\
 & 3 &  & 10 & \textbf{9} &		0.0	& 0.0 & \textbf{9} & 1.5 & \textbf{9} & 9 & 7.7  \\
 & 4 &  & 12 & \textbf{10} &		0.0	& 0.0 & \textbf{10} & 12.1 & \textbf{10} & 10 & 21.3  \\
 & 5 &  & 12 & \textbf{10} &		0.0	& 0.0 & \textbf{10} & 7.1 & \textbf{10} & 10 & 22.0 \\ \hline
15 & 1 & 4 & 13 & \textbf{11} &	0.0	& 0.3 & \textbf{11} & 51.7 & \textbf{11} & 11 & 55.4  \\
 & 2 &  & 12 & \textbf{11} &		0.0	& 1.2 & \textbf{11} & 37.7 & \textbf{11} & 11 & 35.4  \\
 & 3 &  & 11 & \textbf{10} &		0.0	& 1.1 & \textbf{10} & 19.0  & \textbf{10} & 10 & 36.6 \\
 & 4 &  & 13 & \textbf{11} &		0.0	& 1.6 & \textbf{11} & 104.8  & \textbf{11} & 11 & 77.7\\
 & 5 &  & 14 & \textbf{11} &		0.0	& 1.6 & \textbf{11} & 80.2  & \textbf{11} & 11 & 107.0  \\ \hline
15 & 1 & 5 & 14 & \textbf{11} &	0.0	& 8.8 & 14 & TL & \textbf{11} & 11 & 129.1  \\ 
 & 2 &  & 13 & \textbf{11} &		0.0	& 9.1 & \textbf{11} & 239.4 & \textbf{11} & 11 & 60.0 \\
 & 3 &  & 12 & \textbf{11} &		0.0	& 0.7 & \textbf{11} & 162.9  & \textbf{11} & 11 & 72.2\\
 & 4 &  & 14 & \textbf{12} &		0.0	& 2.4 & {12} & TL & 0 & 11 & TL \\
 & 5 &  & 15 & \textbf{12} & 	0.0	& 5.7 & 12 & TL & \textbf{12} & 12 & 210.0 \\ \hline \hline
\end{tabular}}

\end{table}

	\begin{table}[H]
	\caption{Comparison of the approaches on benchmark set \textsc{Random} for $n=20$. } \label{tab:results-random-n-20}
	\centering
	\footnotesize{
		\begin{tabular}{rrrrrrrlllll}
			\hline \hline
			\multicolumn{2}{c}{Instance} &
			\multicolumn{1}{c}{$k$} &
			\multicolumn{1}{c}{\textsc{Greedy}}   &
			\multicolumn{3}{c}{\textsc{VNS}}      &
			\multicolumn{2}{c}{\textsc{Ilp}}      &
			\multicolumn{3}{c}{\textsc{Benders}} \\
   		\cmidrule(rr){1-2}   \cmidrule(r){4-4} \cmidrule(rrr){5-7} \cmidrule(rr){8-9} \cmidrule(rrr){10-12}	 
   		
			$n$ & \textit{index} & & obj & $\overline{obj}$ &  $\sigma[\%]$ & $\overline{t}_{best}[s]$ & obj & $t[s]$ & obj & dual &  $t[s]$ \\ \hline
20 & 1 & 2 & 12 & \textbf{11} &	0.0	& 0.0 & \textbf{11} & 1.5 & \textbf{11} & 11 & 5.5 \\
 & 2 &  & 13 & \textbf{12} & 	0.0	& 0.3 & \textbf{12} & 0.5 & \textbf{12} & 12 & 14.0 \\
 & 3 &  & 12 & \textbf{11} &		0.0	& 0.0 & \textbf{11} & 0.9 & \textbf{11} & 11 & 6.3 \\
 & 4 &  & \textbf{11} & \textbf{11} &		0.0	& 0.0 & \textbf{11} & 0.4 & \textbf{11} & 11 & 5.8 \\
 & 5 &  & \textbf{12} & \textbf{12} &		0.0	& 0.0 & \textbf{12} & 1.4 & \textbf{12} & 12 & 9.3 \\ \hline
20 & 1 & 3 & 14 & \textbf{12} &	0.0	& 4.9 & \textbf{12} & 7.7 & \textbf{12} & 12 & 24.9 \\
 & 2 &  & 16 & \textbf{13} & 	0.0	& 1.0 & \textbf{13} & 17.0 & \textbf{13} & 13 & 109.8 \\
 & 3 &  & 14 & \textbf{13} & 	0.0	& 0.1 & \textbf{13} & 46.2 & \textbf{13} & 13 & 66.2 \\
 & 4 &  & 14 & \textbf{13} & 	0.0	& 0.4 & \textbf{13} & 14.7 & \textbf{13} & 13 & 38.0 \\
 & 5 &  & 15 & \textbf{13} & 	0.0	& 1.5 & \textbf{13} & 41.1 & \textbf{13} & 13 & 62.4 \\ \hline
20 & 1 & 4 & 16 & \textbf{14} &	0.0	& 4.7 & \textbf{14} & TL & \textbf{14} & 14 & 224.2 \\
 & 2 &  & 18 & \textbf{14} & 	0.0	& 17.2 & 18 & TL & 0 & 14 & TL \\
 & 3 &  & 15 & \textbf{13} & 	0.0	& 103.5 & 23 & TL & \textbf{13} & 13 & 89.5 \\
 & 4 &  & 16 & \textbf{14} & 	0.0	& 12.6 & 18 & TL & \textbf{14} & 14 & 111.2 \\
 & 5 &  & 17 & \textbf{14} & 	0.0	& 105.6 & 22 & TL & 0 & 14 & TL \\ \hline
20 & 1 & 5 & 18 & \textbf{15} &	0.0	& 7.4 & - & - & 0 & 14 & TL \\ 
 & 2 &  & 19 & \textbf{15} & 	2.1	& 91.1 & - & - & 0 & 14 & TL \\
 & 3 &  & 16 & \textbf{14} & 	0.0	&43.7 & - & - & 0 & 14 & TL \\
 & 4 &  & 17 & \textbf{14} & 	0.0	&74.7 & - & - & \textbf{14} & 14 & 276.2 \\
 & 5 &  & 19 & \textbf{15} &		0.0	& 66.8 & - & - & 0 & 14 & TL \\ \hline \hline
\end{tabular}}

\end{table}

	\begin{table}
	\caption{Comparison of the approaches on benchmark set \textsc{Random} for $n=45$. } \label{tab:results-random-n-45}
	\centering
	\footnotesize{
		\begin{tabular}{rrrrrrrlllll}
			\hline \hline
			\multicolumn{2}{c}{Instance} &
			\multicolumn{1}{c}{$k$} &
			\multicolumn{1}{c}{\textsc{Greedy}}   &
			\multicolumn{3}{c}{\textsc{VNS}}      &
			\multicolumn{2}{c}{\textsc{Ilp}}      &
			\multicolumn{3}{c}{\textsc{Benders}} \\ 
			
			\cmidrule(rr){1-2}   \cmidrule(r){4-4} \cmidrule(rrr){5-7} \cmidrule(rr){8-9} \cmidrule(rrr){10-12}	 
					
			$n$ & \textit{index} & & obj & $\overline{obj}$ &  $\sigma[\%]$ & $\overline{t}_{best}[s]$ & obj & $t[s]$ & obj & dual &  $t[s]$ \\ \hline
45 & 1 & 2 & 28   & {26.5} &	2.0	& 98.4& \textbf{26} & 14.5 & \textbf{26} & 26 & 240.3\\
 & 2 & & 28       & \textbf{26} & 	0.0	& 9.1 & \textbf{26} & 19.5 & \textbf{26} & 26 & 311.3  \\
 & 3 & & 28       & \textbf{26} & 	0.0	& 7.6 & \textbf{26} & 35.4 & \textbf{26} & 26 & 326.0 \\
 & 4 & & 28       & \textbf{27} & 	0.0	& 8.5 & \textbf{27} & 21.0  & 0 & 27 & TL \\
 & 5 & & 27       & \textbf{26} & 	0.0	& 27.6 & \textbf{26} & 10.3 & \textbf{26} & 26 & 225.9  \\ \hline
45 & 1 & 3 & 35   & \textbf{30} &	1.6	& 166.9 & - & - & 0 & 27 & TL \\
 & 2 & & 34       & \textbf{29.3} & 	1.6	& 283.4 & - & - & 0 & 26 & TL \\
 & 3 & & 35       & \textbf{29.8} & 	2.1	&243.7 & - & - & 0 & 27 & TL \\
 & 4 & & 35       & \textbf{29.8} & 	2.1	& 226.9 & - & - & 0 & 27 & TL \\
 & 5 & & 34       & \textbf{29.4} & 	1.8	& 276.4 & - & - & 0 & 26 & TL \\ \hline
45 & 1 & 4 & 40   & \textbf{32.9} & 	3.0	& 191.5 & - & - & 0 & 28 & TL \\
 & 2 & & 39       & \textbf{32.1} &  1.8	& 259.6 & - & - & 0 & 25 & TL \\
 & 3 & & 40       & \textbf{33.5} & 	2.5	& 228.6 & - & - & 0 & 26 & TL \\
 & 4 & & 40       & \textbf{32.9} & 	3.0	& 276.1 & - & - & 0 & 26 & TL \\
 & 5 & & 38       & \textbf{32.3} & 	2.5	& 208.7 & - & - & 0 & 27 & TL \\
 & 1 &  & 43      & \textbf{36.4} & 	1.4	& 333.4 & - & - & 0 & 28 & TL \\ \hline
45 & 2 & 5& 42    & \textbf{36.0} & 	1.3	& 363.0 & - & - & 0 & 26 & TL \\
 & 3 & & 43       & \textbf{37.9} & 	2.0	& 244.6 & - & - & 0 & 26 & TL \\
 & 4 & & 43       & \textbf{37.1} & 	2.4	& 278.8 & - & - & 0 & 26 & TL \\
 & 5 & & 42       & \textbf{35.9} & 	2.4	&348.6 & - & - & 0 & 26 & TL \\ \hline \hline
\end{tabular}}
\end{table}

	\section{Complete  results for benchmark set \textsc{Wireless}}\label{app: wireless-remaining-results}
	\begin{table}[H]
	\centering
\footnotesize{
\begin{tabular}{rrrrrrlllll}
 \hline \hline
			\multicolumn{2}{c}{Instance} &
			\multicolumn{1}{c}{\textsc{Greedy}}   &
			\multicolumn{3}{c}{\textsc{VNS}}      &
			\multicolumn{2}{c}{\textsc{Ilp}}      &
			\multicolumn{3}{c}{\textsc{Benders}} \\
			
			\cmidrule(rr){1-2}   \cmidrule(r){3-3} \cmidrule(rrr){4-6} \cmidrule(rr){7-8} \cmidrule(rrr){9-11}

$n$ & $R$ &  obj & $\overline{obj}$ & $\sigma[\%]$ & $\overline{t}_{best}[s]$ &  obj & $t[s]$  & obj & dual & $t[s]$ \\ \hline
20 & 0.3 &  13 & \textbf{12} &	0.0	& 0.8 & \textbf{12} & 0.25 & \textbf{12} & 12 & 6.5 \\
 & 0.4 &  10 & \textbf{9} & 	0.0	&0.2 & \textbf{9} & 1.16 & \textbf{9} & 9 & 7.8 \\
 & 0.5 &  6 & \textbf{5} &		0.0	& 0.1 & \textbf{5} & 1.14 & \textbf{5} & 5 & 6.6 \\
 & 0.6 &  \textbf{3} & \textbf{3}	& 0.0 & 0.0 & \textbf{3} & 0.21 & \textbf{3} & 3 & 2.6 \\ \hline

50 & 0.3 &  15 & \textbf{13} & 	0.0	& 3.8 & - & - & \textbf{13} & 13 & 196.2 \\
 & 0.4 &  11 & \textbf{10} & 	0.0	&7.7 & - & - & \textbf{10} & 10 & 177.6 \\
 & 0.5 &  8 & \textbf{6} & 		0.0	&4.0 & - & - & \textbf{6} & 6 & 64.0 \\
 & 0.6 &  6 & \textbf{5} &	0.0		& 2.1 & - & - & \textbf{5} & 5 & 100.1 \\ \hline

100 & 0.3 &  18 & \textbf{13} & 3.0	&239.4 & - & - & 0 & 13 & TL \\
 & 0.4 &  13 & \textbf{10} & 	0.0	&24.2 & - & - & 0 & 10 & TL \\
 & 0.5 &  \textbf{9} & \textbf{9} &	0.0	& 0.0 & - & - & 0 & 7 & TL \\
 & 0.6 &  7 & \textbf{6} & 		0.0&21.5 & - & - & - & - & - \\  \hline \hline
\end{tabular}}
\caption{Comparison of the approaches on benchmark set \textsc{Wireless} for $k=2$. } \label{tab:results-wireless-k-2}
\end{table}
	\begin{table}[H]
	\centering
	\footnotesize{
		\begin{tabular}{rrrrrrlllll}
			\hline \hline
			\multicolumn{2}{c}{Instance} &
			\multicolumn{1}{c}{\textsc{Greedy}}   &
			\multicolumn{3}{c}{\textsc{VNS}}      &
			\multicolumn{2}{c}{\textsc{Ilp}}      &
			\multicolumn{3}{c}{\textsc{Benders}} \\
			
			\cmidrule(rr){1-2}   \cmidrule(r){3-3} \cmidrule(rrr){4-6} \cmidrule(rr){7-8} \cmidrule(rrr){9-11}	 
			
			$n$ & $R$ &  obj & $\overline{obj}$ & $\sigma[\%]$ & $\overline{t}_{best}[s]$ &  obj & $t[s]$  & obj & dual & $t[s]$ \\ \hline
20 & 0.3 &  19 & \textbf{17}	&0.0 & 45.4 & - & - & 0 & 16 & TL \\
 & 0.4 &  16 & \textbf{14.3}	& 3.4& 85.6 & - & - & 0 & 14 & TL \\
 & 0.5 &  9 & \textbf{8} 		& 0.0& 10.3 & - & - & 0 & 8 & TL \\
 & 0.6 &  6 & \textbf{6}		& 0.0 & 0.0 & - & - & 6 & 6 & 187.0 \\ \hline
 
50 & 0.3 &  \textbf{24} & \textbf{24}&0.0 & 0.0 & - & - & 0 & 16 & TL \\
 & 0.4 &  18 & \textbf{17.9}	& 0.0 & 18.9 & - & - & 0 & 12 & TL \\
 & 0.5 &  14 & \textbf{13.7}	& 3.4 &  46.9 & - & - & 0 & 9 & TL \\ 
 & 0.6 &  9 & \textbf{7} 		& 0.0 &  136.8 & - & - & 0 & 7 & TL \\ \hline

100 & 0.3 &  \textbf{33}  & \textbf{33}& 0.0 & 0.0 & - & - & - & - & - \\
 & 0.4 &  25 & \textbf{24.6}	& 1.8 & 182.5 & - & - & - & - & - \\
 & 0.5 &  \textbf{18} & \textbf{18} & 3.5 & 0.0 & - & - & - & - & - \\
 & 0.6 &  \textbf{12} & \textbf{12}	& 0.0 & 0.0 & - & - & - & - & - \\ \hline \hline
\end{tabular}}
\caption{Comparison of the approaches on benchmark set \textsc{Wireless} for $k=5$. } \label{tab:results-wireless-k-5}
\end{table}

	\bibliographystyle{abbrv}
	\bibliography{bib}

\end{document}